 \documentclass[pmlr,twocolumn,10pt,table]{jmlr} 

\usepackage{booktabs}

\usepackage{siunitx}

\usepackage[switch]{lineno}

\newcommand{\equal}[1]{{\hypersetup{linkcolor=black}\thanks{#1}}}

\theorembodyfont{\upshape}
\theoremheaderfont{\scshape}
\theorempostheader{:}
\theoremsep{\newline}

\usepackage[english]{babel}
\usepackage[autostyle, english=american]{csquotes}
\MakeOuterQuote{"}

\definecolor{myblue}{RGB}{0,114,178}
\definecolor{myorange}{RGB}{213,94,0}

\usepackage{booktabs}
\usepackage{caption}
\usepackage{multirow}
\usepackage{newfloat}
\usepackage{listings}
\usepackage{dsfont}
\usepackage{arydshln}
\usepackage{comment}
\usepackage{mathtools}
\usepackage{lipsum}
\usepackage{algorithm}

\let\classAND\AND
\let\AND\relax
\usepackage{algorithmic}

\let\AND\classAND
\AtBeginEnvironment{algorithmic}{\let\AND\algoAND}

\usepackage{bm}
\usepackage{bbm}

\usepackage{adjustbox}
\usepackage{makecell}

\DeclarePairedDelimiter\autobracket{(}{)}
\newcommand{\br}[1]{\autobracket*{#1}}

\newcommand{\bx}{\bm{x}}
\newcommand{\bth}{\bm{\theta}}

\renewcommand{\Pr}{\text{Pr}}

\usepackage{dcolumn}
\newcolumntype{d}[1]{D{.}{.}{#1}}
\DeclareUnicodeCharacter{2212}{-}

\def\R{\mathbbm{R}}
\def\ie{{\em i.e.},\ }
\def\eg{{\em e.g.},\ }
\long\def\comment#1{}

\jmlrvolume{297}
\jmlryear{2025}
\jmlrworkshop{Machine Learning for Health (ML4H) 2025} 

 \title[MENSA: A Multi-Event Network for Survival Analysis]{MENSA: A Multi-Event Network for Survival Analysis with Trajectory-based Likelihood Estimation}

\author{%
\Name{Christian Marius Lillelund}\textsuperscript{1,2}\equal{Corresponding author.} \Email{clillelund@ualberta.ca}
\AND
\Name{Ali Hossein Gharari Foomani}\textsuperscript{1}
\Email{hosseing@ualberta.ca}
\AND
\Name{Weijie Sun}\textsuperscript{1} \Email{weijie2@ualberta.ca}
\AND
\Name{Shi-ang Qi}\textsuperscript{1} \Email{shiang@ualberta.ca}
\AND
\Name{Russell Greiner}\textsuperscript{1,3} \Email{rgreiner@ualberta.ca}
\AND
\Name{{The Pooled Resource Open-Access ALS Clinical Trials Consortium (PRO-ACT)}}\textsuperscript{4} \\
\addr $^{1}$Department of Computing Science, University of Alberta, Edmonton, Canada \\
\addr $^{2}$Department of Electrical and Computer Engineering, Aarhus University, Aarhus, Denmark\\
\addr $^{3}$Alberta Machine Intelligence Institute, Edmonton, Canada\\
\addr $^{4}$PRO-ACT Consortium$\dagger$
}

\begin{document}

\maketitle

\begin{abstract}
Most existing time-to-event methods focus on either single-event or competing-risks settings, leaving multi-event scenarios relatively underexplored. In many healthcare applications, for example, a patient may experience multiple clinical events, that can be non-exclusive and semi-competing. A common workaround is to train independent single-event models for such multi-event problems, but this approach fails to exploit dependencies and shared structures across events. To overcome these limitations, we propose MENSA (Multi-Event Network for Survival Analysis), a deep learning model that jointly learns flexible time-to-event distributions for multiple events, whether competing or co-occurring. In addition, we introduce a novel trajectory-based likelihood term that captures the temporal ordering between events. Across four multi-event datasets, MENSA improves predictive performance over many state-of-the-art baselines. Source code is available at \url{https://github.com/thecml/mensa}.
\end{abstract}


\paragraph*{Data and Code Availability} All datasets used in this study are publicly available. Appendix \ref{app:datasets_and_preprocessing} describes how to obtain these datasets and preprocess them accordingly.

\paragraph*{Institutional Review Board (IRB)}
This study uses only publicly available datasets and does not involve any interaction with human subjects or access to identifiable private information. Therefore, IRB approval was not required.

\section{Introduction}
\label{sec:introduction}

Many survival models estimate the probability that an event of interest (typically death) will occur for an individual at some future time. This can involve: a \emph{single event} of interest, where we want to model the time until one specific event~\citep[Ch. 11]{gareth_introduction_2021}, \emph{competing risks}, where the occurrence of one event excludes other events from occurring~\citep{fine1999proportional}, or \emph{multiple events}, where several events may occur in some order~\citep{andersen_multi_2002}. As an example of the latter, a patient with a chronic condition may experience several medical events over the course of their disease.

In current practice, multi-event problems are often approached by training multiple single-event models~\citep{armstrong_aging_2014, solomon_angiotensin_2017}. However, this approach (1) fails to exploit shared covariate patterns, (2) ignores temporal or causal dependencies between events, and (3) becomes computationally inefficient as the number of events increases. Other methods mitigate these issues through shared frailty terms~\citep{jiang_semi_2017}, multi-task regularization~\citep{li_multi-task_2016, wang_multi-task_2017}, or joint survival modeling~\citep{hsieh_quantile_2018}, but these rely on restrictive assumptions about frailty, proportional hazards, or a fixed survival distribution. Recent work has explored structured transition models~\citep{groha2020general}, hierarchical time-scale modeling~\citep{tjandra_hierarchical_2021}, latent ordinary differential equations (ODEs) for estimating event-specific hazard curves~\citep{moon2022survlatent}, and transition models based on Monte Carlo sampling~\citep{Rossman2022}. These offer partial solutions but typically assume predefined state transitions, scale poorly to long time horizons, cannot handle multiple events, or depend on sampling to predict an individual survival distributions (ISD)~\citep{haider_effective_2020}. To date, most research remains confined to single-event or competing-risks scenarios. Motivated by these limitations, we introduce a new and scalable deep learning model for multi-event survival analysis, called MENSA (Multi-Event Network for Survival Analysis). Our main contributions are:

\begin{itemize}
\item We propose a multi-state survival model based on the framework by \citet{andersen_multi_2002}, where each event corresponds to a transition between two states (\eg from healthy to sick). This enables prediction of how long a patient will remain in a particular state, or when the patient will experience one of multiple events. Each event-specific marginal distribution is modeled as a mixture of Weibull distributions, relaxing the proportional hazards assumption. Also, this architecture naturally extends to single-event and competing-risks settings.
\item We introduce a training objective that combines the standard negative log-likelihood (NLL) loss with an additional trajectory-based term. This new term encodes known temporal relationships between events, guiding the model to predict ISDs that align with prior knowledge.
\item We show empirically that MENSA offers solid discrimination and calibration performance, improving over many baseline models. MENSA is also more computationally efficient than several baselines in the multi-event setting. Two ablation studies show that training MENSA jointly across all events leads to a notable increase in the local concordance index (C-index), reflecting better within-patient event ordering, and that using the proposed trajectory-based likelihood term further improves this metric.
\end{itemize}

\section{Related work}

\citet{andersen_multi_2002} proposed a model for multi-state processes (for example, disease progression leading to relapse and eventually death), capturing sequences of events. This framework has since become a fundamental tool for multi-state analysis~\citep{gorfine2025overview}. Each transition between states (events) is modeled with its own hazard function, allowing state-specific effects, and naturally incorporates competing risks, recurrent events, and absorbing states (\eg death). However, individual event probabilities must be derived from transition probabilities across states, so the event time cannot be computed directly. Moreover, the model assumes a linear relationship between covariates and the transition-specific hazard functions.

\citet{groha2020general} proposed an ODE-based neural network for multi-state survival analysis that incorporates a predefined state-transition structure. This structure is encoded using a binary transition matrix, indicating which transitions are allowed. The model then learns time-varying transition intensities by solving Kolmogorov forward equations, but only for those transitions permitted by the predefined graph. A key limitation of this approach is that it enforces a fixed sequence of events -- for example, once cancer occurs, the model does not allow subsequent heart failure and vice versa. This framework requires practitioners to specify the full state-transition graph in advance, which can be restrictive when the event order is uncertain or variable.

\citet{tjandra_hierarchical_2021} proposed a hierarchical multi-event model that predicts event occurrences across multiple time scales, using coarse estimates (\eg monthly) to guide finer predictions (\eg daily). This approach decomposes long-horizon tasks into simpler intermediate steps, progressively refining predictions from broad to fine temporal resolutions. However, it has two main limitations. First, the hierarchical structure incurs significant parameter overhead and scales poorly with longer durations. Second, the method is optimized solely for discriminative loss. Strong discrimination performance does not necessarily translate into accurate time-to-event predictions, and vice versa~\citep{qi_effective_2023}.

\citet{moon2022survlatent} proposed an ODE-based model for multi-event analysis. This model maps temporal features into continuous latent trajectories and estimates cause-specific hazard functions at discrete time intervals without relying on parametric hazard assumptions. However, the model is only designed for competing risks, supports at most two events, and assumes access to time-varying features. It cannot be directly applied on datasets with only static (baseline) covariates, which remain the most common in survival analysis. Moreover, while latent ODE-based models offer high flexibility, they face practical limitations. As noted by \citet[p. 66]{chen_2024_introduction}, these models are typically slower to train and more prone to numerical instability than parametric alternatives.

\citet{Rossman2022} introduced a multi-state survival model, \emph{PyMSM}, which estimates transition-specific hazard functions using time-dependent covariates and Monte Carlo sampling. PyMSM is well-suited for settings where individuals move through a sequence of well-defined states (\eg from illness to recovery to death), and covariates can be updated at each transition. However, its reliance on Monte Carlo sampling introduces two key limitations. First, to estimate an ISD for a patient, the model must sample many possible trajectories, which can be computationally expensive. Second, since the model only approximates the ISDs, it lacks closed-form survival or density functions, which prevents analytical comparisons (\eg hazard ratios) or mathematical guarantees (\eg proportional hazards). 

\section{The proposed method}

\subsection{Problem formulation}

We consider a survival dataset with multiple events and right-censoring, consisting of \emph{N} time-to-event triplets, denoted as $\mathcal{D} = \left\lbrace \br{\bm{x}^{(i)}, \bm{t}^{(i)}, \bm{\delta}^{(i)}} \right\rbrace_{i=1}^N$, where $\bm{x}^{(i)} \in \mathbb{R}^{d}$ denotes a $d$-dimensional vector of features (covariates), $\bm{t}^{(i)} \in \{1,2,\ldots,t_{\text{max}}\}^{K}$ is a vector of observed times for the $i$-th instance across $K$ events, and $\bm{\delta}^{(i)} \in \{0,1\}^{K}$ is a vector of event indicators for each event\footnote{In our terminology, we refer to \emph{competing-risks} problems as those in which only one event can be observed per instance, and the event time is effectively a scalar: $t^{(i)}$. In contrast, \emph{multi-event} problems involve multiple events per instance, and the event time is instead represented as a vector: $\bm{t}^{(i)}$. Cases involving \emph{semi-competing risks} -- where one event precludes another but not vice versa -- fall under our definition of multi-event problems.}. We consider each instance to have (potential) event times, $e_{k}^{(i)} \in \{1,2,\ldots,t_{\text{max}}\}$, and censoring times, $c_{k}^{(i)} \in \{1,2,\ldots,t_{\text{max}}\}$. We then have $t_{k}^{(i)} = e_{k}^{(i)}$ if $\delta_{k}^{(i)} = 1$, meaning that the event was observed, otherwise, $t_{k}^{(i)} = c_{k}^{(i)}$ if $\delta_{k}^{(i)} = 0$, meaning that the event was right-censored.

\begin{figure}[!ht]
\centering
\includegraphics[width=1\columnwidth, trim=10 0 10 0, clip]{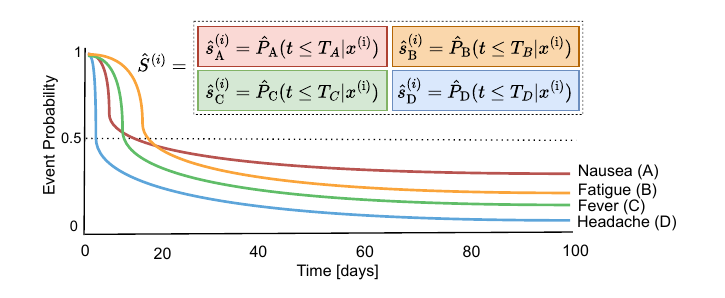}
\caption{Example of a multi-event prediction.}
\label{fig:mensa_multi_event}
\end{figure}

Now, as a practical example, consider a drug trial over $t \in [0, 100]$ days (see Figure \ref{fig:mensa_multi_event}), where a patient $i$ with features $\bm{x}^{(i)}$ can experience four side effects: nausea (A), fatigue (B), fever (C), or headache (D). Based on $\bm{x}^{(i)}$, we want to estimate the so-called survival function \( S_E(t) = \Pr(t < T_E) \)\footnote{We use \( T_E \) to denote the time of event, and \( T_C \) to denote the time of censoring. The observed time is then \( T = \min(T_E, T_C) \).} for each of these possible events (side effects), which is the probability that the patient has not yet experienced the event by \( t \geq 0 \). By estimating this function for every event, we obtain a matrix of survival functions, $\hat{\bm{S}}^{(i)} = [\bm{s}_{1}^{(i)},\bm{s}_{2}^{(i)},\ldots,\bm{s}_{K}^{(i)}]$, for the $i$-th patient, where $K$ is the number of events. Finally, we can estimate the patient's likely time to each event as the point where each curve crosses the 50\% (median) line.

\subsection{Mixture model architecture}

\begin{figure*}[!ht]
\centering
\includegraphics[width=1\textwidth, trim=10 10 100 10, clip]{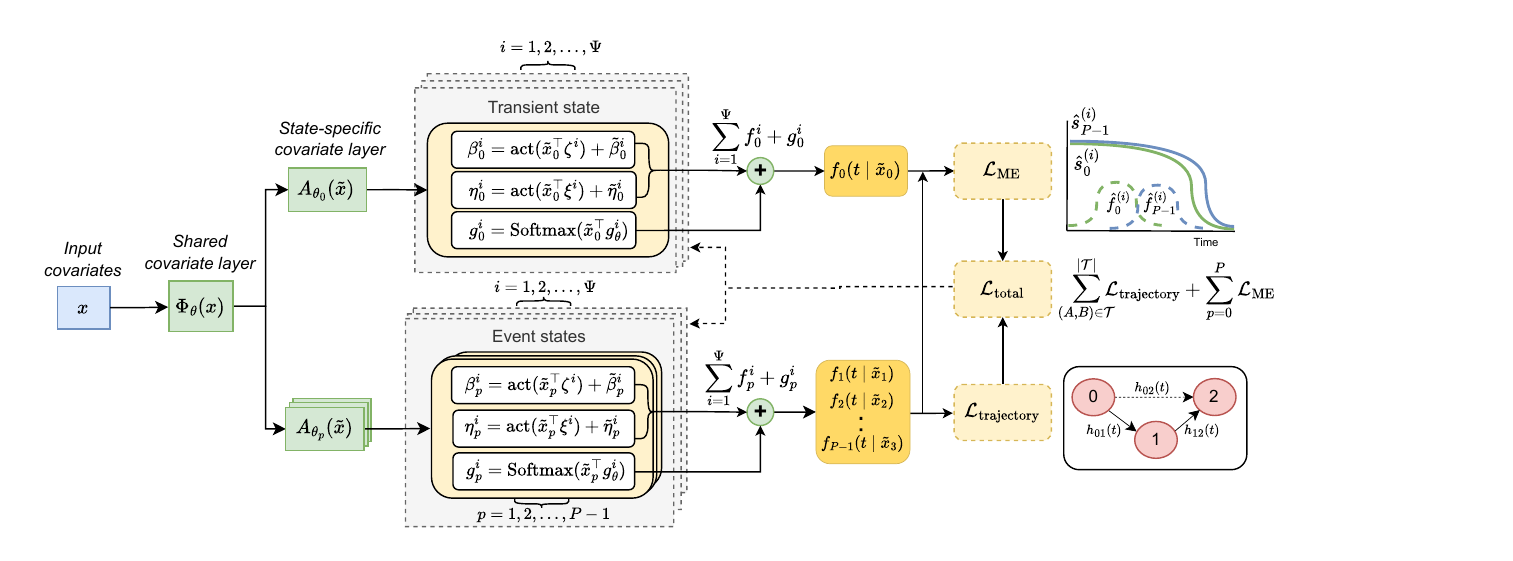}
\caption{Overview of the MENSA architecture. The model first maps input covariates through a shared multilayer perceptron (MLP) layer, producing a common representation \(\tilde{\bm{x}}\). This is passed through event-specific covariate layers, each producing \(\tilde{\bm{x}}_{p}\) for state \(p \in \mathcal{P}\). Each adapted representation then parameterizes a mixture of Weibull distributions to predict a time-to-event density function, \( f_p(t) \), capturing the time of transition from the transient state (state 0) to each event state. The model is trained using a combined loss consisting of a multi-event likelihood, \(\mathcal{L}_{\text{ME}}\), which captures how well the model fits event times across all events, and a trajectory likelihood, \(\mathcal{L}_{\text{trajectory}}\), which enforces known temporal ordering constraints between pairs of events.}
\label{fig:mensa_architecture}
\end{figure*}

We propose a multi-state architecture (see Figure~\ref{fig:mensa_architecture}) in which events are modeled as transitions between discrete states -- \eg from \emph{healthy} to \emph{sick}, or from \emph{alive} to \emph{dead} -- based on the framework by \citet{andersen_multi_2002}, and further described in Appendix \ref{app:multi_state_analysis}. Let \( \mathcal{P} = \{0, 1, \ldots, P{-}1\} \) denote the finite set of states, where state 0 is the initial (event-free) state. Each event \( k \in \{1, \ldots, K\} \) corresponds to a transition into a state \( p_k \in \mathcal{P} \setminus \{0\} \). For example, if event \( k \) represents a \emph{relapse}, then \( p_k \) denotes the absorbing state for \emph{relapse}. This mapping naturally supports scenarios with sequential or co-occurring events -- \eg first becoming ill and then later dying.

For each instance \( i \), we estimate a matrix of survival distributions \( \bm{S}^{(i)} \): one distribution for each state \( p \in \mathcal{P} \), including the initial event-free state (\( p = 0 \)). The marginal density function \( f_p(t) \) for state $p$, which represents the probability of entering state \( p \) at time point $t$, is modeled as a convex combination of \( \Psi \in \mathbb{N}_{+} \) Weibull distributions. This formulation offers two key advantages. First, it relaxes the proportional hazards assumption on the underlying marginal survival distribution (proof in Appendix~\ref{app:proof_of_proportional_hazards}), allowing the hazard function to vary over time. Second, it leverages the closed-form probability density function (PDF) and cumulative distribution function (CDF) of the Weibull distribution, respectively:
\begin{align}
f(t; \beta, \eta) &= \frac{\eta}{\beta} \left( \frac{t}{\beta} \right)^{\eta - 1} \exp\left( -\left( \frac{t}{\beta} \right)^{\eta} \right), \\
F(t; \beta, \eta) &= 1 - \exp\left( -\left( \frac{t}{\beta} \right)^{\eta} \right),
\end{align}
where \(\beta > 0\) is the scale and \(\eta > 0\) the shape parameter. We parameterize each such Weibull distribution by two covariate-dependent transformations, that is, $h_p(\bm{x}^{(i)}) = A_{\theta,p}\!\big(\Phi_{\theta}(\bm{x}^{(i)})\big)$, such that:
\begin{align}
\log\beta_{\psi,p}(\bm{x}^{(i)})
&= \tilde{\beta}_{\psi,p}
   + \mathrm{act}\!\big( h_p(\bm{x}^{(i)})^{\!\top}\zeta_{\psi,p} \big),
\label{eq:log_beta}\\
\log\eta_{\psi,p}(\bm{x}^{(i)})
&= \tilde{\eta}_{\psi,p}
   + \mathrm{act}\!\big( h_p(\bm{x}^{(i)})^{\!\top}\xi_{\psi,p} \big),
\label{eq:log_eta}
\end{align}

where $\Phi_\theta(\cdot)$ is a shared covariate transformation, $A_{\theta,p}(\cdot)$ is a state-specific adapter applied to its output, $\mathrm{act}(\cdot)$ denotes a non-linear activation function, $\psi \in \{1,2,\ldots,\Psi\}$ indexes the Weibull mixture distributions, and $\zeta_{\psi,p}, \xi_{\psi,p} \in \mathbb{R}^d$ are learnable weight vectors associated with the scale and shape parameters, respectively. The terms $\tilde{\beta}_{\psi,p}$ and $\tilde{\eta}_{\psi,p}$ are learnable biases that initialize the log-scale and log-shape, respectively, for each state $p$. We define $\bm{\theta}_{\psi,p}(\bm{x}^{(i)}) = \big(\beta_{\psi,p}(\bm{x}^{(i)}), \eta_{\psi,p}(\bm{x}^{(i)})\big)$, and the corresponding distribution PDF and CDF for transitioning into state $p$ at time $t$ given covariates $\bm{x}^{(i)}$ are then given by:
\begin{align}
f_p(t;\bm{x}^{(i)})
&= \sum_{\psi=1}^{\Psi} g_{\psi,p}(\bm{x}^{(i)})\, f_{\psi,p}\!\big(t;\bm{\theta}_{\psi,p}(\bm{x}^{(i)})\big),
\label{eq:mensa_pdf}\\
F_p(t;\bm{x}^{(i)})
&= \sum_{\psi=1}^{\Psi} g_{\psi,p}(\bm{x}^{(i)})\, F_{\psi,p}\!\big(t;\bm{\theta}_{\psi,p}(\bm{x}^{(i)})\big),
\label{eq:mensa_cdf}
\end{align}
where $g_{\psi,p}(\bm{x}^{(i)})$ are normalized mixture weights produced by a state-specific gating network, \ie a softmax over linear outputs of $h_p(\bm{x}^{(i)})$. The corresponding survival function is $S_p(t;\bm{x}^{(i)}) = 1 - F_p(t;\bm{x}^{(i)})$.

\subsection{Trajectory-based likelihood}

\begin{table*}[!ht]
\caption{Possible scenarios for two events, $A$ and $B$, with corresponding transition probabilities and log-likelihood contributions. $\delta_A$ and $\delta_B$ indicate whether each event is observed ($1$) or censored ($0$). $P_{0 \rightarrow A}(t_A)$ and $P_{0 \rightarrow B}(t_B)$ denote the probabilities of experiencing events $A$ and $B$ at times $t_A$ and $t_B$, respectively; $S_A(t)$ and $S_B(t)$ are survival functions; and $f_A(t)$ and $f_B(t)$ are probability density functions. If prior knowledge dictates that event $A$ must precede event $B$ when both occur, the joint-event likelihood ($\delta_A = 1$, $\delta_B = 1$) incorporates the probability of surviving event $B$ at the time of event $A$, as highlighted in bold.}
\label{tab:possible_event_scenarios}
\centering
\begin{tabular}{lllll}
\toprule
\# & $\delta_A$ & $\delta_B$ & Transition probability & Log-likelihood \\
\midrule
1 & 0 & 0 & $1 - (P_{0 \rightarrow A}(t_A)+  P_{0 \rightarrow B}(t_B))$ & $S_A(t_A) + S_B(t_B) $ \\
2 & 1 & 0 & $P_{0 \rightarrow A}(t_A)$ & $f_A(t_A) + S_B(t_B) $ \\
3 & 0 & 1 & $P_{0 \rightarrow B}(t_B)$ & $S_A(t_A) + f_B(t_B) $ \\
4 & 1 & 1 & $P_{0 \rightarrow A}(t_A)*P_{0 \rightarrow B}(t_B)$ & $f_A(t_A) + f_B(t_B) + \boldsymbol{S_B(t_A)}$ \\
\bottomrule
\end{tabular}

\end{table*}

To estimate \eqref{eq:mensa_pdf} and \eqref{eq:mensa_cdf}, we extend the standard survival likelihood (defined in Appendix~\ref{app:survival_likelihood}) to our multi-state architecture -- where each instance $i$ may experience multiple events. However, rather than indexing events directly, we express the likelihood in terms of state transitions. Specifically, we define one indicator and time pair \( (\delta_p^{(i)}, t_p^{(i)}) \) per state \( p \in \mathcal{P} \), where each event \( k \) corresponds to a transition into state \( p \). The indicator \( \delta_p^{(i)} \) denotes whether a transition into state \( p \) was observed, and \( t_p^{(i)} \) is the corresponding observed or censored transition time. The likelihood includes the state's density \( f_{p}(t_p^{(i)} \mid \bm{x}^{(i)}) \), if observed (\( \delta_p^{(i)} = 1 \)), and the survival function \( S_{p}(t_p^{(i)} \mid \bm{x}^{(i)}) \), if censored (\( \delta_p^{(i)} = 0 \)).

To account for the imbalance between state transitions, we introduce non-negative weights, \( w_p \), computed from the inverse frequency of transitions into each state using the training set, normalized so that \( \sum_{p \in \mathcal{P}} w_p = |\mathcal{P}| \). Assuming conditional independence between state transitions\footnote{Conditional independence means that, given the covariates, the transition to one state is assumed to be independent of transitions to other states. Therefore, any relationship between states is assumed explained by the covariates.}, we can then define our multi-event likelihood as:
\begin{align}\label{eq:me_event_likelihood}
\mathcal{L}_{\text{ME}}(\mathcal{D})
= \sum_{i=1}^N \sum_{p \in \mathcal{P}} w_p \Big[
  \delta^{(i)}_{p}\, \log f_{p}\!\left(t^{(i)}_{p} \mid \bm{x}^{(i)}\right) \\ \notag
 +\quad (1-\delta^{(i)}_{p})\, \log S_{p}\!\left(t^{(i)}_{p} \mid \bm{x}^{(i)}\right)
\Big],
\end{align}
where \( f_{p}(\cdot) \) and \( S_{p}(\cdot) \) denote the conditional density and survival function for state \( p \), respectively.

To guide the model, we incorporate prior knowledge of temporal event ordering when available. As an example, consider the illness-death model~\citep{fix1951simple}, where patients start out healthy (state 0), become ill (event A), and may eventually die (event B). This means there are four possible scenarios (see Table \ref{tab:possible_event_scenarios}): (1) neither event occurred (both censored), (2) only A occurred (B censored), (3) only B occurred (A censored), and (4) both events occurred, with A before B. We assume that if such semi-competing events are observed, one must come before the other -- \eg death cannot occur before illness. To incorporate this constraint into the optimization, we define an auxiliary likelihood term that encourages the model to assign a high survival probability for event \( B \) at the time event \( A \) occurs, \ie maximize \( S_B(t_A) \). This ensures the model assigns a low probability to B when A has occurred, effectively handling semi-competing risks.

Let \( \mathcal{T} = \left\{ (A, B) \mid A, B \in \mathcal{K}, \, \text{A occurs before B} \right\} \) denote the set of known event orderings, and for each instance \( i \), we consider all uncensored event pairs \( (A, B) \in \mathcal{T} \). The trajectory likelihood is computed by summing the survival probability of event $B$ at the time of event $A$, thus $S_B(T_A \mid \bm{x}^{(i)})$, weighted by the event indicators $\delta_A^{(i)}$ and $\delta_B^{(i)}$, which equal 1 if events $A$ and $B$, respectively, have occurred for instance $i$, and 0 otherwise. The likelihood is then:
\begin{align}
\mathcal{L}_{\text{trajectory}}(\mathcal{D}) &= \sum_{i=1}^N \sum_{(A,B)\in \mathcal{T}}
\delta_A^{(i)} \delta_B^{(i)} \, \log S_B\!\big(T_A \mid \bm{x}^{(i)}\big), \label{eq:traj_likelihood}
\end{align}
where \( \mathcal{T} \) is the set of trajectories. Finally, we have:
\begin{align}\label{eq:total_likelihood}
\mathcal{L}_{\text{total}}(\mathcal{D}) &= (1 - \lambda)\frac{1}{N} \mathcal{L}_{\text{ME}}(\mathcal{D}) 
+ \lambda \cdot \frac{1}{N} \mathcal{L}_{\text{trajectory}}(\mathcal{D}),
\end{align}
where $\lambda$ is a weighting hyperparameter that controls the trade-off between the two likelihoods. The training algorithm can be found in Appendix \ref{app:implementation_details}. 

\section{Experiments}

\subsection{Experimental setup}

\textbf{Datasets.} We use five datasets to test the effectiveness of our method. Table \ref{tab:datasets} in Appendix \ref{app:experimental_details} summarizes the dataset statistics, and Appendix \ref{app:datasets_and_preprocessing} contains details of preprocessing steps, and flow charts of event trajectories. These datasets cover single-event, competing-risks and multi-event settings.

\textbf{Baselines.} We compare 11 baseline survival models: Cox proportional hazards (CoxPH)~\citep{cox_regression_1972}, CoxNet~\citep{simon_regularization_2011}, Weibull AFT~\citep{klein2006survival}, GBSA~\citep{Ridgeway1999}, RSF~\citep{ishwaran_random_2008}, NFG\footnote{The Neural Fine Gray (NFG) model is used exclusively for the competing-risks experiments, see Appendix \ref{app:competing_risks_results}.}~\citep{jeanselme23a}, DeepSurv~\citep{katzman_deepsurv_2018}, DeepHit~\citep{lee_deephit_2018}, Hierarchical~\citep{tjandra_hierarchical_2021}, MTLR~\citep{NIPS2011_1019c809, kim_deep-cr_2021}, and DSM~\citep{nagpal_deep_2021}. Appendix \ref{app:baselines} provides implementation details.

\textbf{Hyperparameters.} To ensure a fair comparison, all models -- including baselines -- were tuned using Bayesian optimization~\citep{snoek_practical_2012} over ten iterations. For each model and dataset, we selected the hyperparameters that achieved the highest Harrell's C-index~\citep{harrell1996} on the validation set. Appendix \ref{app:hyperparameter_settings} lists the chosen parameters.

\textbf{Evaluation metrics.} To evaluate the discrimination performance, we report the global C-index (CI\textsubscript{G})~\citep{tjandra_hierarchical_2021}, which measures how well the model can discriminate instances within an event, and the local C-index (CI\textsubscript{L})~\citep{tjandra_hierarchical_2021}, which measures how well the model can discriminate events within an instance, and the time-dependent AUC, which measures discrimination at a fixed time point. We calculate the AUC by computing it at the 25th, 50th, and 75th percentiles of observed event times in the test set and then averaging these values. To evaluate calibration performance, we report the mean absolute error (MAE) using a margin loss (mMAE)~\citep{haider_effective_2020}, the integrated Brier score (IBS)~\citep{graf1999assessment} and the distribution calibration (D-calibration) score~\citep{haider_effective_2020}. Appendix \ref{app:evaluation_metrics} gives more details on the metrics.

\subsection{Implementation details}

Following recent literature~\citep{katzman_deepsurv_2018, nagpal_deep_2021, 
lillelund_efficient_2024}, we use a multilayer perceptron (MLP) for the backbone network architecture. This transforms the raw input features into a learned representation, which is then used to parameterize the Weibull distributions. On top of this shared representation, we add small state-specific MLPs to fine-tune representations for each prediction task. We use a single hidden layer, the ReLU6 activation function for the MLP, and a mixture of $\Psi$ Weibull distributions. Adam is utilized for optimization~\citep{KingBa15}. Our model is implemented in Pytorch~\citep{Paszke2019}. See Appendix \ref{app:implementation_details} for further details.

In each of 10 experiments, we follow~\citet{Sechidis2011} and split each dataset into stratified training (70\%), validation (10\%), and test (20\%) sets with random seeds 0–9. After splitting, we impute missing values using the sample mean for continuous covariates and the mode for categorical covariates, based on the training set. We apply a $z$-score normalization to numerical features and encode categorical features using one-hot encoding. We train MENSA with early stopping on the validation set with a patience of 20 epochs. Concerning reporting, results are first computed separately for each random seed and event. Metrics are then averaged across events within each seed, producing one value per seed per model–dataset combination. We report the mean and standard deviation of these per-seed averages.

\subsection{Results and takeaways}

\begin{table*}[!ht]
\centering
\caption{Discrimination performance (mean $\pm$ SD.) between different models on multi-event benchmark datasets, averaged over 10 experiments. For all metrics, higher ($\uparrow$) is better. Model types are abbreviated as SE = single event, CR = competing risks, and ME = multiple events. Bold indicates the top-3 models. $\dagger$ indicates models that are significantly and practically worse than MENSA, according to paired $t$-tests with Holm correction ($\alpha=0.05$) and a minimum effect of interest (MEI) threshold of 1.0.}
\label{tab:multi_event_discrimination}
\resizebox{\textwidth}{!}{%
\begin{tabular}{@{}llcccccccccccc@{}}
\toprule
\multirow{2}{*}{\centering Model} & \multirow{2}{*}{Type} &
\multicolumn{3}{c}{MIMIC-IV ($K=3$)} &
\multicolumn{3}{c}{Rotterdam ($K=3$)} &
\multicolumn{3}{c}{PRO-ACT ($K=4$)} &
\multicolumn{3}{c}{EBMT ($K=5$)} \\
& &
CI\textsubscript{G} & CI\textsubscript{L} & AUC &
CI\textsubscript{G} & CI\textsubscript{L} & AUC &
CI\textsubscript{G} & CI\textsubscript{L} & AUC &
CI\textsubscript{G} & CI\textsubscript{L} & AUC \\
\midrule
CoxPH & SE & 70.9\text{\tiny{$\pm$0.52}} & 90.3\text{\tiny{$\pm$0.48}} & 74.3\text{\tiny{$\pm$0.71}} & 69.7\text{\tiny{$\pm$1.16}} & 83.6\text{\tiny{$\pm$2.09}}$\dagger$ & 75.1\text{\tiny{$\pm$1.48}} & 70.6\text{\tiny{$\pm$0.47}} & \textbf{76.5}\text{\tiny{$\pm$0.55}} & 76.7\text{\tiny{$\pm$0.91}} & 54.9\text{\tiny{$\pm$1.04}} & 66.3\text{\tiny{$\pm$1.16}}$\dagger$ & 54.6\text{\tiny{$\pm$1.10}} \\
CoxNet & SE & 71.0\text{\tiny{$\pm$0.53}} & 90.4\text{\tiny{$\pm$0.45}} & 74.4\text{\tiny{$\pm$0.71}} & 69.8\text{\tiny{$\pm$1.16}} & 83.8\text{\tiny{$\pm$2.05}}$\dagger$ & 75.1\text{\tiny{$\pm$1.49}} & 70.6\text{\tiny{$\pm$0.46}} & 76.5\text{\tiny{$\pm$0.57}} & 76.7\text{\tiny{$\pm$0.89}} & \textbf{55.1}\text{\tiny{$\pm$1.12}} & 67.2\text{\tiny{$\pm$1.13}}$\dagger$ & 54.5\text{\tiny{$\pm$1.17}} \\
Weibull & SE & 71.2\text{\tiny{$\pm$0.54}} & \textbf{90.8}\text{\tiny{$\pm$0.43}} & 74.8\text{\tiny{$\pm$0.70}} & 69.7\text{\tiny{$\pm$1.18}} & 87.9\text{\tiny{$\pm$1.99}}$\dagger$ & 75.1\text{\tiny{$\pm$1.45}} & 70.7\text{\tiny{$\pm$0.49}} & \textbf{77.1}\text{\tiny{$\pm$0.57}} & 76.7\text{\tiny{$\pm$0.90}} & 54.9\text{\tiny{$\pm$1.36}} & \textbf{68.0}\text{\tiny{$\pm$0.97}} & 54.2\text{\tiny{$\pm$1.35}} \\
GBSA & SE & \textbf{71.9}\text{\tiny{$\pm$0.43}} & \textbf{91.4}\text{\tiny{$\pm$0.51}} & \textbf{75.2}\text{\tiny{$\pm$0.53}} & \textbf{75.3}\text{\tiny{$\pm$1.62}} & \textbf{90.5}\text{\tiny{$\pm$1.05}} & 71.2\text{\tiny{$\pm$2.02}}$\dagger$ & \textbf{72.0}\text{\tiny{$\pm$0.31}} & 75.6\text{\tiny{$\pm$0.66}} & \textbf{77.6}\text{\tiny{$\pm$0.44}} & \textbf{55.3}\text{\tiny{$\pm$0.93}} & 67.2\text{\tiny{$\pm$0.82}}$\dagger$ & \textbf{55.4}\text{\tiny{$\pm$1.26}} \\
RSF & SE & 69.1\text{\tiny{$\pm$0.48}}$\dagger$ & \textbf{90.8}\text{\tiny{$\pm$0.40}} & 74.6\text{\tiny{$\pm$0.56}} & \textbf{70.7}\text{\tiny{$\pm$1.15}} & 84.9\text{\tiny{$\pm$2.19}}$\dagger$ & \textbf{77.3}\text{\tiny{$\pm$1.74}} & \textbf{71.1}\text{\tiny{$\pm$0.51}} & 75.6\text{\tiny{$\pm$0.83}} & \textbf{77.5}\text{\tiny{$\pm$0.60}} & 55.0\text{\tiny{$\pm$0.75}} & 64.6\text{\tiny{$\pm$0.92}}$\dagger$ & \textbf{55.6}\text{\tiny{$\pm$1.07}} \\
MTLR & SE & 70.2\text{\tiny{$\pm$0.50}}$\dagger$ & 88.3\text{\tiny{$\pm$0.95}}$\dagger$ & 74.4\text{\tiny{$\pm$0.71}} & 69.8\text{\tiny{$\pm$1.16}} & 82.8\text{\tiny{$\pm$2.37}}$\dagger$ & \textbf{75.7}\text{\tiny{$\pm$1.85}} & 70.7\text{\tiny{$\pm$0.53}} & \textbf{76.5}\text{\tiny{$\pm$0.53}} & 77.4\text{\tiny{$\pm$0.83}} & 53.6\text{\tiny{$\pm$1.04}} & 61.6\text{\tiny{$\pm$2.38}}$\dagger$ & 55.4\text{\tiny{$\pm$1.19}} \\
DeepSurv & SE & \textbf{71.8}\text{\tiny{$\pm$0.55}} & 89.9\text{\tiny{$\pm$0.49}} & \textbf{75.8}\text{\tiny{$\pm$0.68}} & 70.0\text{\tiny{$\pm$1.24}} & 84.6\text{\tiny{$\pm$2.93}}$\dagger$ & 75.4\text{\tiny{$\pm$1.75}} & 71.0\text{\tiny{$\pm$0.59}} & 76.0\text{\tiny{$\pm$0.57}} & 76.7\text{\tiny{$\pm$1.00}} & \textbf{55.1}\text{\tiny{$\pm$1.02}} & 67.6\text{\tiny{$\pm$0.82}} & 54.9\text{\tiny{$\pm$1.22}} \\
DeepHit & CR & 70.9\text{\tiny{$\pm$0.75}} & 63.1\text{\tiny{$\pm$9.74}}$\dagger$ & 75.1\text{\tiny{$\pm$0.94}} & 66.5\text{\tiny{$\pm$1.09}}$\dagger$ & \textbf{90.2}\text{\tiny{$\pm$1.03}} & 74.2\text{\tiny{$\pm$1.50}} & 64.9\text{\tiny{$\pm$1.15}}$\dagger$ & 65.3\text{\tiny{$\pm$3.35}}$\dagger$ & 73.4\text{\tiny{$\pm$0.73}}$\dagger$ & 52.4\text{\tiny{$\pm$1.29}}$\dagger$ & 60.4\text{\tiny{$\pm$2.76}}$\dagger$ & 55.2\text{\tiny{$\pm$1.46}} \\
DSM & CR & 70.5\text{\tiny{$\pm$0.60}} & 88.1\text{\tiny{$\pm$1.37}}$\dagger$ & 74.5\text{\tiny{$\pm$0.83}} & \textbf{70.6}\text{\tiny{$\pm$1.26}} & 86.0\text{\tiny{$\pm$2.81}}$\dagger$ & \textbf{76.4}\text{\tiny{$\pm$1.68}} & 70.8\text{\tiny{$\pm$0.68}} & 76.0\text{\tiny{$\pm$0.68}} & \textbf{77.4}\text{\tiny{$\pm$0.76}} & 54.8\text{\tiny{$\pm$1.00}} & \textbf{68.0}\text{\tiny{$\pm$0.99}} & 55.3\text{\tiny{$\pm$1.36}} \\
Hierarch. & ME & 65.2\text{\tiny{$\pm$1.67}}$\dagger$ & 85.3\text{\tiny{$\pm$1.73}}$\dagger$ & 48.7\text{\tiny{$\pm$0.51}}$\dagger$ & 67.8\text{\tiny{$\pm$2.06}} & 55.7\text{\tiny{$\pm$8.11}}$\dagger$ & 50.0\text{\tiny{$\pm$0.05}}$\dagger$ & 70.8\text{\tiny{$\pm$0.60}} & 73.6\text{\tiny{$\pm$1.42}}$\dagger$ & 57.3\text{\tiny{$\pm$0.89}}$\dagger$ & 54.7\text{\tiny{$\pm$1.33}} & 44.5\text{\tiny{$\pm$1.12}}$\dagger$ & 53.5\text{\tiny{$\pm$1.03}}$\dagger$ \\
MENSA (Ours) & ME & \textbf{71.4}\text{\tiny{$\pm$0.29}} & 90.3\text{\tiny{$\pm$0.54}} & \textbf{75.2}\text{\tiny{$\pm$0.41}} & 69.4\text{\tiny{$\pm$1.20}} & \textbf{90.1}\text{\tiny{$\pm$1.55}} & 73.8\text{\tiny{$\pm$2.31}} & \textbf{70.8}\text{\tiny{$\pm$0.45}} & 76.4\text{\tiny{$\pm$0.61}} & 77.2\text{\tiny{$\pm$0.60}} & 55.1\text{\tiny{$\pm$1.37}} & \textbf{68.8}\text{\tiny{$\pm$0.66}} & \textbf{55.7}\text{\tiny{$\pm$1.62}} \\
\bottomrule
\end{tabular}
}
\end{table*}

\begin{table*}[!ht]
\centering
\caption{Calibration performance (mean $\pm$ SD.) between different models on four multi-event benchmark datasets, averaged over 10 experiments. For IBS and mMAE, lower ($\downarrow$) is better. For D-calibration, higher ($\uparrow$) is better. D-calibration counts the number of experiments in which the respective model was D-calibrated. Bold indicates the top-3 models. $\dagger$ indicates models that are significantly and practically worse than MENSA, based on paired $t$-tests with Holm correction ($\alpha=0.05$) and a predefined MEI of 1.0 for IBS and mMAE.}
\label{tab:multi_event_calibration}
\resizebox{\textwidth}{!}{%
\begin{tabular}{@{}llcccccccccccc@{}}
\toprule
\multirow{2}{*}{\centering Model} & \multirow{2}{*}{Type} &
\multicolumn{3}{c}{MIMIC-IV ($K=3$)} &
\multicolumn{3}{c}{Rotterdam ($K=3$)} &
\multicolumn{3}{c}{PRO-ACT ($K=4$)} &
\multicolumn{3}{c}{EBMT ($K=5$)} \\
& &
IBS & mMAE & D-Cal &
IBS & mMAE & D-Cal &
IBS & mMAE & D-Cal &
IBS & mMAE & D-Cal \\
\midrule
CoxPH & SE & 15.3\text{\tiny{$\pm$0.51}} & 10.8\text{\tiny{$\pm$5.77}} & 30/30 & 19.6\text{\tiny{$\pm$1.53}} & 23.2\text{\tiny{$\pm$1.15}}$\dagger$ & 20/20 & 16.6\text{\tiny{$\pm$0.39}} & 129.0\text{\tiny{$\pm$3.31}}$\dagger$ & 36/40 & 20.4\text{\tiny{$\pm$0.68}} & 48.6\text{\tiny{$\pm$0.71}} & 49/50 \\
CoxNet & SE & 15.3\text{\tiny{$\pm$0.52}} & 8.6\text{\tiny{$\pm$0.16}} & 30/30 & 19.6\text{\tiny{$\pm$1.54}} & 23.0\text{\tiny{$\pm$1.11}} & 20/20 & 16.6\text{\tiny{$\pm$0.40}} & 128.6\text{\tiny{$\pm$3.23}}$\dagger$ & 36/40 & 20.3\text{\tiny{$\pm$0.68}} & \textbf{48.3}\text{\tiny{$\pm$0.70}} & 49/50 \\
Weibull & SE & 15.6\text{\tiny{$\pm$0.42}} & 11.3\text{\tiny{$\pm$0.14}}$\dagger$ & 30/30 & 19.2\text{\tiny{$\pm$1.64}} & 23.1\text{\tiny{$\pm$0.78}} & 15/20 & 16.7\text{\tiny{$\pm$0.35}} & 220.7\text{\tiny{$\pm$11.24}}$\dagger$ & 36/40 & \textbf{20.1}\text{\tiny{$\pm$0.98}} & 59.4\text{\tiny{$\pm$0.86}} & 18/50 \\
GBSA & SE & 16.7\text{\tiny{$\pm$0.63}}$\dagger$ & \textbf{6.9}\text{\tiny{$\pm$0.07}} & 29/30 & 19.3\text{\tiny{$\pm$1.44}} & 24.1\text{\tiny{$\pm$0.74}}$\dagger$ & 20/20 & 16.7\text{\tiny{$\pm$0.42}} & \textbf{120.8}\text{\tiny{$\pm$2.14}} & 35/40 & 20.4\text{\tiny{$\pm$0.75}} & \textbf{44.7}\text{\tiny{$\pm$0.44}} & 50/50 \\
RSF & SE & 16.6\text{\tiny{$\pm$0.49}}$\dagger$ & \textbf{7.5}\text{\tiny{$\pm$0.08}} & 27/30 & \textbf{18.6}\text{\tiny{$\pm$1.29}} & \textbf{22.6}\text{\tiny{$\pm$1.05}} & 18/20 & \textbf{16.3}\text{\tiny{$\pm$0.41}} & 170.1\text{\tiny{$\pm$22.18}}$\dagger$ & 39/40 & \textbf{18.9}\text{\tiny{$\pm$1.25}} & 68.9\text{\tiny{$\pm$1.00}} & 50/50 \\
MTLR & SE & \textbf{14.9}\text{\tiny{$\pm$0.52}} & 9.7\text{\tiny{$\pm$0.15}}$\dagger$ & 30/30 & \textbf{18.1}\text{\tiny{$\pm$1.65}} & 23.0\text{\tiny{$\pm$0.72}} & 19/20 & \textbf{16.4}\text{\tiny{$\pm$0.42}} & \textbf{126.3}\text{\tiny{$\pm$4.56}} & 40/40 & \textbf{18.9}\text{\tiny{$\pm$1.25}} & 68.9\text{\tiny{$\pm$1.00}}$\dagger$ & 50/50 \\
DeepSurv & SE & \textbf{14.9}\text{\tiny{$\pm$0.51}} & 9.1\text{\tiny{$\pm$0.22}} & 30/30 & 19.4\text{\tiny{$\pm$1.47}} & \textbf{22.7}\text{\tiny{$\pm$1.02}} & 18/20 & \textbf{16.3}\text{\tiny{$\pm$0.42}} & 139.1\text{\tiny{$\pm$5.27}}$\dagger$ & 37/40 & 20.5\text{\tiny{$\pm$0.73}} & \textbf{45.8}\text{\tiny{$\pm$0.63}} & 49/50 \\
DeepHit & CR & 32.7\text{\tiny{$\pm$0.98}}$\dagger$ & 20.0\text{\tiny{$\pm$0.31}}$\dagger$ & 0/30 & 21.4\text{\tiny{$\pm$2.64}}$\dagger$ & 27.3\text{\tiny{$\pm$0.79}}$\dagger$ & 18/20 & 25.6\text{\tiny{$\pm$1.21}}$\dagger$ & 196.5\text{\tiny{$\pm$4.00}}$\dagger$ & 0/40 & 21.1\text{\tiny{$\pm$3.99}} & 74.3\text{\tiny{$\pm$0.84}}$\dagger$ & 50/50 \\
DSM & CR & 16.0\text{\tiny{$\pm$0.47}}$\dagger$ & 17.0\text{\tiny{$\pm$1.92}}$\dagger$ & 22/30 & \textbf{18.3}\text{\tiny{$\pm$1.54}} & 23.2\text{\tiny{$\pm$0.76}}$\dagger$ & 15/20 & 16.5\text{\tiny{$\pm$0.35}} & 231.5\text{\tiny{$\pm$15.51}}$\dagger$ & 38/40 & 20.7\text{\tiny{$\pm$1.46}} & 59.6\text{\tiny{$\pm$0.82}} & 17/50 \\
Hierarch. & ME & 51.0\text{\tiny{$\pm$1.16}}$\dagger$ & 101.9\text{\tiny{$\pm$233.61}}$\dagger$ & 0/30 & 56.0\text{\tiny{$\pm$5.09}}$\dagger$ & 43.5\text{\tiny{$\pm$0.76}}$\dagger$ & 0/20 & 53.9\text{\tiny{$\pm$1.86}}$\dagger$ & 280.5\text{\tiny{$\pm$5.47}}$\dagger$ & 0/40 & 71.8\text{\tiny{$\pm$7.84}}$\dagger$ & 92.2\text{\tiny{$\pm$0.80}}$\dagger$ & 0/50 \\
MENSA (Ours) & ME & \textbf{15.1}\text{\tiny{$\pm$0.59}} & \textbf{8.3}\text{\tiny{$\pm$0.09}} & 21/30 & 18.8\text{\tiny{$\pm$1.59}} & \textbf{22.6}\text{\tiny{$\pm$0.68}} & 18/20 & 16.6\text{\tiny{$\pm$0.41}} & \textbf{124.1}\text{\tiny{$\pm$2.14}} & 39/40 & 22.0\text{\tiny{$\pm$2.33}} & 63.1\text{\tiny{$\pm$2.88}} & 10/50 \\
\bottomrule
\end{tabular}
}
\end{table*}

\begin{figure*}[!ht]
\centering
\includegraphics[width=1\textwidth]{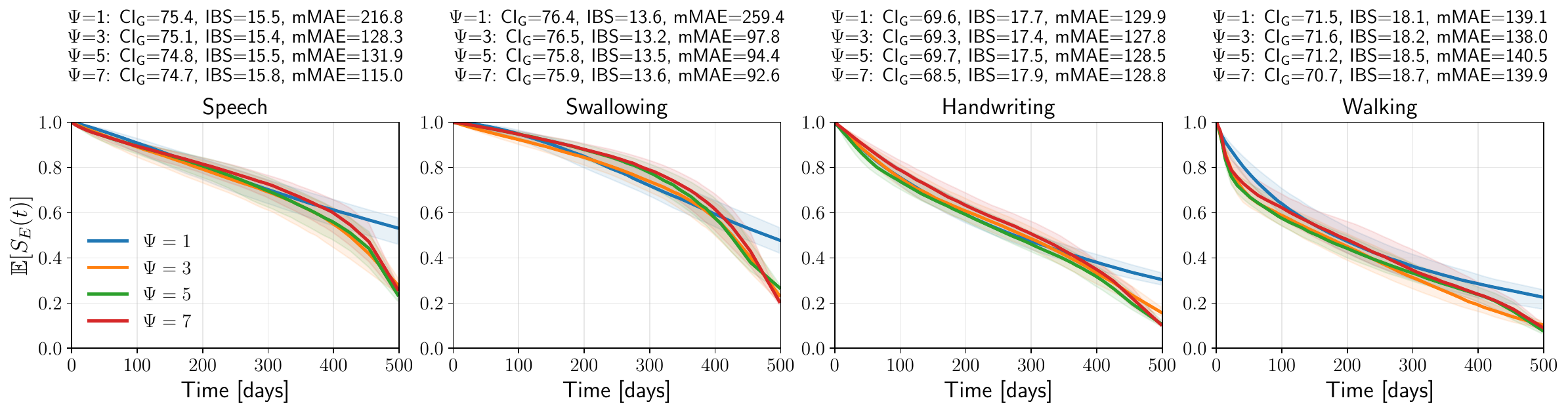}
\caption{Population-level summaries of predicted survival functions across the PRO-ACT test set for each event. Each panel shows the median expected survival $\mathbb{E}[S_E(t)]$ with narrow (40–60th percentile) variability bands. Different colors correspond to MENSA models trained with varying numbers of Weibull distributions $\Psi \in \{1, 3, 5, 7\}$. The inset boxes report event-specific performance metrics (Global C-index, IBS, mMAE).}
\label{fig:proact_pop_survival}
\end{figure*}

\begin{table*}[!ht]
\centering
\caption{Ablation study comparing joint vs. independent training of MENSA on two multi-event datasets. Results are averaged over 10 experiments, with changes shown relative ($\%$) to the baseline.}
\label{tab:ab_joint_vs_independent}
\resizebox{1\textwidth}{!}{%
\begin{tabular}{@{}clllllll@{}}
\toprule
Dataset & Training & CI\textsubscript{G} & CI\textsubscript{L} & AUC & IBS & mMAE & D-Cal \\ 
\midrule
\multirow{2}{*}{\makecell{Rotterdam}}
& Joint & 69.45$\pm$1.25 & 90.24$\pm$1.36 & 73.88$\pm$2.33 & 18.78$\pm$1.59 & 22.62$\pm$0.71 & 18/20 \\
& Indep. & 69.82$\pm$1.31 \textcolor{myblue}{(+0.53)} & 87.81$\pm$1.74 \textcolor{myorange}{(-2.69)} & 74.57$\pm$2.10 \textcolor{myblue}{(+0.93)} & 18.84$\pm$1.67 \textcolor{myorange}{(+0.32)} & 22.48$\pm$0.67 \textcolor{myblue}{(-0.63)} & 20/20 \\
\cmidrule(lr){1-1}
\multirow{2}{*}{\makecell{PRO-ACT}}
& Joint & 70.87$\pm$0.50 & 76.07$\pm$0.51 & 77.12$\pm$1.20 & 16.60$\pm$0.52 & 124.37$\pm$3.44 & 36/40 \\
& Indep. & 70.98$\pm$0.64 \textcolor{myblue}{(+0.15)} & 75.92$\pm$0.88 \textcolor{myorange}{(-0.19)} & 77.29$\pm$0.75 \textcolor{myblue}{(+0.23)} & 16.54$\pm$0.42 \textcolor{myblue}{(-0.34)} & 122.78$\pm$2.33 \textcolor{myblue}{(-1.27)} & 38/40 \\
\bottomrule
\end{tabular}
}
\end{table*}

\begin{table*}[!ht]
\centering
\caption{Ablation study comparing training MENSA with/without the trajectory likelihood on two multi-event datasets. Results are averaged over 10 experiments, with changes shown relative ($\%$) to the baseline.}
\label{tab:ab_trajectory_based_likelihood}
\resizebox{1\textwidth}{!}{%
\begin{tabular}{@{}clllllll@{}}
\toprule
Dataset & $\mathcal{L}_{trajectory}$ & CI\textsubscript{G} & CI\textsubscript{L} & AUC & IBS & mMAE & D-Cal \\ 
\midrule
\multirow{2}{*}{\makecell{Rotterdam}}
& With & 69.45$\pm$1.25 & 90.24$\pm$1.36 & 73.88$\pm$2.33 & 18.78$\pm$1.59 & 22.62$\pm$0.71 & 18/20 \\
& Without & 69.54$\pm$1.22 \textcolor{myblue}{(+0.13)} & 89.94$\pm$1.45 \textcolor{myorange}{(-0.34)} & 74.16$\pm$2.09 \textcolor{myblue}{(+0.38)} & 18.79$\pm$1.72 \textcolor{myorange}{(+0.09)} & 22.58$\pm$0.71 \textcolor{myblue}{(-0.18)} & 18/20 \\
\cmidrule(lr){1-1}
\multirow{2}{*}{\makecell{EBMT}}
& With & 55.26$\pm$1.24 & 68.73$\pm$0.56 & 55.66$\pm$1.47 & 22.13$\pm$2.30 & 62.98$\pm$2.97 & 9/50 \\
& Without & 54.93$\pm$1.15 \textcolor{myorange}{(-0.6)} & 68.39$\pm$0.56 \textcolor{myorange}{(-0.5)} & 55.53$\pm$1.43 \textcolor{myorange}{(-0.24)} & 22.36$\pm$2.59 \textcolor{myorange}{(+1.04)} & 63.76$\pm$3.90 \textcolor{myorange}{(+1.25)} & 14/50 \\
\bottomrule
\end{tabular}
}
\end{table*}

Here, we report results on the multi-event datasets. Concerning the single-event and competing-risks settings, these results are presented in Appendix \ref{app:additional_results}.

\textbf{Overall performance.} Table \ref{tab:multi_event_discrimination} presents the discrimination results of our proposed model and existing baselines on four multi-event datasets. Models marked with $^{\dagger}$ in the tables are significantly and practically worse than MENSA (details of the statistical testing and minimum effect of interest (MEI) thresholds are provided in Appendix \ref{app:stat_tests}). Overall, MENSA outperforms many deep learning baselines in discrimination tasks across multiple datasets and performs comparably to strong tree-based methods such as GBSA and RSF. Unlike DeepHit or Hierarchical, MENSA avoids unstable behavior (large variance or metric collapse). For these multi-event datasets, we also assess local discrimination performance using the local C-index, which captures the accuracy of within-patient event orderings and reflects the added value of a multi-event model. In cases where MENSA outperforms the baselines in local C-index, we attribute the improvement to its multi-state architecture and trajectory-based likelihood term. Table \ref{tab:multi_event_calibration} summarizes calibration performance. MENSA maintains low IBS and mMAE -- often close to or better than classical survival methods (\eg CoxPH, RSF) -- and exhibits comparable D-calibration to other deep learning variants (\eg DeepSurv and DSM). Compared to the Hierarchical model, our approach demonstrates consistently superior performance across all datasets.

\textbf{Impact of number of distributions.} Using functional decline prediction in patients with amyotrophic lateral sclerosis (ALS) as a test case (PRO-ACT dataset; four events: \emph{Speaking}, \emph{Swallowing}, \emph{Handwriting}, and \emph{Walking}; see Appendix~\ref{app:datasets_and_preprocessing}), we train MENSA to estimate event-specific population survival curves. Each curve represents the median survival (\ie event) probability $\mathbb{E}[S_E(t)]$ across all $N=670$ patients under a given mixture configuration $\Psi$. Setting $\Psi = 1$ means that a single distribution models the time-to-event distribution for each event. Increasing $\Psi$ introduces additional distributions, allowing the model to represent more complex and non-proportional risk structures. As shown in Figure~\ref{fig:proact_pop_survival}, higher mixture complexity shifts the population survival downward across all events by the end of follow-up (about 500 days), reflecting a more pessimistic risk estimate. Quantitatively, this added flexibility lowers prediction error (mMAE: 183$\rightarrow$118 days) and slightly improves calibration (IBS: 0.158$\rightarrow$0.155) while maintaining stable discrimination (Global CI: $\approx$0.73). These results suggest that moderate mixture complexity ($\Psi \approx 3-5$) better captures disease progression in ALS without overfitting, improving time-to-event accuracy at the population level.

\textbf{Computational analysis.} We analyze the computational complexity of our method and several deep baselines\footnote{We report computational metrics only for the deep learning models; FLOPs and parameter counts are not well-defined or directly comparable for classical methods.}. For fairness, we configure all models with a default architecture (see Appendix \ref{app:hyperparameter_settings}). The full results are presented in Appendix \ref{app:computational_analysis}. In the multi-event setting, all baselines (except the Hierarchical model) must be trained $K$ times, whereas MENSA handles all events in a single forward pass. On a (hypothetical) dataset with 100 features and four events, MENSA requires 1.4$\times$ fewer parameters than DeepSurv, 2.5$\times$ fewer than DeepHit, 3.2$\times$ fewer than MTLR, 1.5$\times$ fewer than DSM, and 13$\times$ fewer than the Hierarchical model. Reductions in FLOP follow a similar trend, ranging from 1.5$\times$ (DeepSurv, DSM) and 2.5$\times$ (DeepHit) to 5.7$\times$ (MTLR) and 13$\times$ (Hierarchical).

\textbf{Ablation studies.} We perform two ablation studies evaluating (1) joint versus independent training, and (2) the trajectory-based likelihood term. Table~\ref{tab:ab_joint_vs_independent} shows the results for training events jointly. Here, \emph{jointly} refers to learning all event-specific parameters within a single shared architecture. We see in both datasets that the global metrics (\eg global C-index, AUC and mMAE) increase slightly when training independently, but the local C-index (within-patient event ordering) drops significantly in Rotterdam (-2.69\% decrease) and PRO-ACT (-0.19\% decrease). Table \ref{tab:ab_trajectory_based_likelihood} presents the results for the trajectory term. In Rotterdam, it improves the local C-index with only small shifts in the global C-index, AUC and mMAE. In EBMT, the trajectory term consistently improves performance.

\section{Discussion}

\subsection{Limitations}

This work has limitations. First, MENSA assumes conditionally independent censoring, which may be violated if relevant covariates that influence both censoring and event times are unobserved. Second, as a parametric model, MENSA relies on predefined event-time distributions. While this choice enhances interpretability and stability, it can reduce flexibility and require larger datasets for reliable parameter estimation. In contrast, semiparametric approaches such as CoxPH and DeepSurv make fewer distributional assumptions and can sometimes generalize better in small-sample settings. Third, the current implementation of MENSA is limited to static baseline covariates and does not yet incorporate time-varying covariates or longitudinal measurements, which are common in many clinical applications.

\subsection{Application}

MENSA provides particular value in clinical contexts where the \emph{local ordering} of events is most relevant. For example, in the PRO-ACT dataset, the goal is to predict the time to different types of functional decline in ALS patients as measured by the ALSFRS-R scale (see Appendix~\ref{app:datasets_and_preprocessing}). A model with a high local C-index can accurately rank event occurrences within a single patient, enabling clinically meaningful questions such as: \emph{Which physical functions am I likely to lose first, and in what order?} This prioritization of local ordering is especially useful in neurodegenerative diseases like ALS, where functional decline follows distinct trajectories. Prior studies have shown, for instance, that bulbar-onset ALS typically affects speech functionality before gait and walking ability~\citep{Feldman2022ALS}. Beyond ALS, multi-event modeling is crucial in many diseases where multiple outcomes unfold over time. In chronic heart failure, for example, patients may experience repeated hospitalizations, arrhythmic events, and eventual cardiac death, each with complex temporal dependencies. In oncology (\eg the SEER cohort), treatment response, relapse, and metastasis form interdependent event pathways that conventional single-event models cannot adequately capture. Similarly, in hematologic and transplant settings (\eg the EBMT cohort), graft-versus-host disease, relapse, and non-relapse mortality compete and co-occur, requiring models that jointly estimate their risks.

Regarding interpretability, practitioners can extract feature importances for each event-specific prediction. Model-agnostic methods tailored to survival analysis, such as SurvLIME~\citep{Kovalev2020SurvLIME} and SurvSHAP~\citep{Krzyzinski2023SurvSHAPt}, can be applied to MENSA's event-specific outputs to quantify the contribution of individual covariates to each predicted event time.

\subsection{Conclusion}

We have presented MENSA, a new method for multi-event survival analysis. MENSA jointly learns flexible time-to-event distributions for multiple events, which may be competing or co-occurring. Across four multi-event survival datasets, MENSA demonstrated solid discrimination and calibration performance, outperforming many deep learning baselines while remaining computationally efficient in the multi-event setting. Compared to the Hierarchical survival model -- the main existing deep learning approach for multi-event survival analysis -- MENSA achieved superior results across all evaluation metrics. Ablation studies further showed that training all events within a shared architecture improves within-patient event ordering, as reflected by higher local C-index values, and that the proposed trajectory-based likelihood term further enhances this metric.

\acks{
$\dagger$Data used in the preparation of this article were obtained from the Pooled Resource Open-Access ALS Clinical Trials (PRO-ACT) Database. As such, the following organizations and individuals within the PRO-ACT Consortium contributed to the design and implementation of the PRO-ACT Database and/or provided data, but did not participate in the analysis of the data or the writing of this report: ALS Therapy Alliance, Cytokinetics, Inc., Amylyx Pharmaceuticals, Inc., Knopp Biosciences, Neuraltus Pharmaceuticals, Inc., Neurological Clinical Research Institute, MGH, Northeast ALS Consortium, Novartis, Orion Corporation, Prize4Life Israel, Regeneron Pharmaceuticals, Inc., Sanofi, Teva Pharmaceutical Industries, Ltd., The ALS Association.

This research received support from the Natural Science and Engineering Research Council of Canada (NSERC), the Canadian Institute for Advanced Research (CIFAR), and the Alberta Machine Intelligence Institute (Amii).
}

\bibliography{references}

\appendix
\onecolumn

\section{Notation}
\label{apd:notation}

Table~\ref{tab:notation} provides a list of notations used in this work.

\begin{table}[!ht]
\centering
\caption{Table of notation.}
\label{tab:notation}
\begin{tabular}{ll}
\toprule
\textbf{Symbol} & \textbf{Definition} \\
\toprule
$c^{(i)}_{p}$ & Censoring time for instance $i$ in state $p$ \\
$e^{(i)}_{p}$ & True event/transition time for instance $i$ in state $p$ \\
$t^{(i)}_{p}$ & Observed time for instance $i$ in state $p$: $\min(e^{(i)}_{p}, c^{(i)}_{p})$ \\
$\delta^{(i)}_{p}$ & Event indicator: $\delta^{(i)}_{p} = \mathbbm{1}[e^{(i)}_{p} \leq c^{(i)}_{p}]$ \\
$\mathcal{D}$ & A time-to-event dataset \\
$N$ & Number of individuals/instances in a dataset \\
$d$ & Number of features/covariates in a dataset \\
$\bx^{(i)}$ & Covariate vector for instance $i$, $\bx^{(i)} \in \mathbb{R}^d$ \\
$h_0(t)$ & Baseline hazard function \\
$h(t \mid \bx^{(i)})$ & Hazard function given covariates $\bx^{(i)}$ \\
$K \in \mathbb{N}_{+}$ & Number of distinct events \\
$P \in \mathbb{N}_{+}$ & Number of distinct states \\
$\mathcal{K}$ & Set of possible events, $|\mathcal{K}| = K$ \\
$\mathcal{P}$ & Set of possible states, $|\mathcal{P}| = P$ \\
$\mathcal{T}$ & Set of event trajectories over $\mathcal{K}$ \\
$\mathcal{L}$ & Loss/objective function used during training \\
$L$ & Number of training epochs \\
$\kappa$ & Learning rate for optimization \\
$\lambda$ & Weighting hyperparameter for the trajectory loss \\
$\Psi \in \mathbb{N}_{+}$ & Number of mixture distributions \\
$G(t) \in \mathcal{P}$ & Latent state process (\eg Markov process) over state space $\mathcal{P}$ \\
$f(\bth, \bx^{(i)})$ & Risk or output function given parameters $\bth$ and input $\bx^{(i)}$ \\
$S_E(t)$ & Event survival function: $S_E(t) = \Pr(T_E > t)$ \\
$f_E(t)$ & Event density function: $f_E(t) = \frac{d}{dt}(1 - S_E(t))$ \\
$F_E(t)$ & Event cumulative distribution function: $F_E(t) = 1 - S_E(t)$ \\
\bottomrule
\end{tabular}
\end{table}

\section{Multi-state analysis}
\label{app:multi_state_analysis}

\begin{figure}[!ht]
\centering
\includegraphics[width=0.45\textwidth]{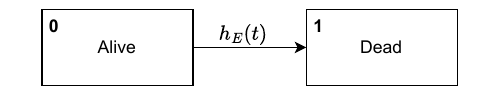}
\caption{A single-event model with a single transition function.}
\label{fig:single_event_model}
\end{figure}

\textbf{Single event.} Survival analysis typically models the time until a single event takes place~\citep[Ch. 11]{gareth_introduction_2021}. We can represent such problems using a two-state model, as in Figure \ref{fig:single_event_model}. 

In this simplest case, there is one transient state (0, \eg "alive") and one absorbing state (1, \eg "dead"). Once an instance enters an absorbing state, no further transitions are possible. The observation for a given instance will in the simplest form consist of a random variable, $T_E$, representing the time from a given origin ($t=0$) to the occurrence of the event. The distribution of $T_E$ can be characterized by the probability distribution function $F_E(t) = \Pr(T_E \leq t)$ or, equivalently, by the survival function $S_E(t) = 1 - F_E(t) = \Pr(T_E > t)$. Note that $S_E(t)$ and $F_E(t)$, respectively, correspond to the probabilities of being in state 0 or 1 at time $t$. In continuous time, the distribution of $T_E$ is commonly characterized by the \emph{hazard rate function}:

\begin{equation}
h_E\br{t} = -d \log S_E(t)/{dt} = \lim_{\Delta t \rightarrow 0} \frac{\Pr\br{t < T_E\leq t+ \Delta t \vert T_E>t}}{\Delta t}\text{,}
\end{equation}
where $S_E(t)$ is the \emph{survival function}:
\begin{equation}
S_E(t) = \exp \left( - \int_0^{t} h_E(u) \, du \right)\text{.}
\end{equation}

Here, $h_E\br{t}$ is the transition intensity from state 0 to 1, \ie the instantaneous risk of going from state 0 to state 1 at time $t$. We can extend this framework to multiple state transitions using \emph{marked point processes}~\citep{andersen_multi_2002}. Let $G(t)$, $t \in \mathbb{T}, \quad \mathbb{T} = [0, t_{\text{max}}]$ denote a stochastic process with a finite state space $\mathcal{P} = \{0, ..., P-1\}$ and with right-continuous sample paths: $G(t+) = G(t)$. The process has an initial distribution $\pi_p(0) = [\Pr(G(0) = p) \mid p \in \mathcal{P}]$. A multi-state process $G(\cdot)$ generates a history $\mathcal{G}_t$, which consists of observations of the process in the interval $[0, t_{\text{max}}]$ -- \eg the patient is healthy at the start of the study ($\mathcal{G}_{0}$), but becomes ill 10 days in ($\mathcal{G}_{10}$), before the study terminates after 100 days ($t_{\text{max}}=100$). Given this history, we can define the so-called transition probabilities between two distinct events by:

\begin{equation}
\Pr_{p \rightarrow q}(s, t) = \text{Pr}(G(t) = q \mid G(s) = p, \mathcal{G}_{s^{-}}),
\end{equation}
for $p,q \in \mathcal{P}, s, t \in \mathbb{T}, s \leq t$ and transition intensities by the derivatives, assuming such exist:

\begin{equation}
h_{p \rightarrow q}(t) = \lim_{\Delta t \to 0} \frac{\Pr_{p \rightarrow q}(t, t + \Delta t)}{\Delta t}.
\end{equation}

Some transition intensities may be zero for all $t$ between two states, \ie $h_{p \rightarrow q}(t) = 0 \; \forall \, t \text{ and } \forall \, p,q \in \mathcal{P}$. A state $p \in \mathcal{P}$ is absorbing if, for all $t \in \mathbb{T}, p \in \mathcal{P}, p \neq q, h_{p \rightarrow q}(t) = 0;$ otherwise $p$ is transient. The state probabilities $\pi_{p}(t) = \text{Pr}(G(t) = p)$ are given by:
\begin{equation}
\pi_{p}(t) = \sum_{q \in \mathcal{P}} \pi^{q}(0) \; \Pr_{q \rightarrow p}(0,t).
\end{equation}

Note that $\Pr_{p \rightarrow q}(\cdot, \cdot)$ and $h_{p \rightarrow q}(\cdot)$ depend on both the probability measure $\text{Pr}(\cdot)$, and on the history.

\begin{figure}[!ht]
\centering
\includegraphics[width=0.4\textwidth]{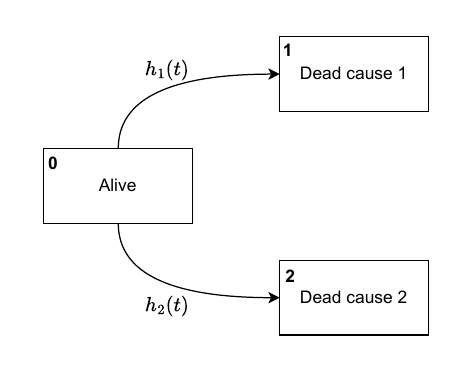}
\caption{A competing-risks model for mortality from two causes.}
\label{fig:competing_risks_model}
\end{figure}

\textbf{Competing risks.} In competing risks, multiple events (\ie absorbing states) exist, but only one event of $k = 1, \dots,  K$ different types can occur for each instance, illustrated in Figure \ref{fig:competing_risks_model}.

In our multi-state framework, a competing-risks model has one transient state, aka state 0, and a number of absorbing states. Entering an absorbing state corresponds to experiencing event \( k \), for \( k \in \{1, \ldots, K\} \). Thus, we have $P$ total states and $K = P-1$ events. In this model, if an instance is censored before experiencing any events, it simply remained in the transient state and is thus event-free. Using our state notation, the transition intensities $h_{p}(t)$ for $p \in \{1, \ldots, P-1\}$ are given by the cause-specific hazard function (omitting state 0):
\begin{equation}
h_p(t) = \lim_{\Delta t \to 0} \frac{{}\Pr(\text{Transition to state}\; p \; \text{at} \; t + \Delta t \mid T_p \geq t)}{\Delta t}\text{,}
\end{equation}
where $T_p$ is the random variable corresponding to the time of the transition to state $p$. There are only edges originating from state 0, the only transient state of the model, \ie $h_{p \rightarrow q}(t) = 0$ for all $p \neq 0$ and all $q$. The transition probabilities are given by the survival function:
\begin{equation}
\Pr_{0 \rightarrow 0}(0, t) = S_0(t) = \Pr(T_0 > t) = \exp \left( - \int_0^t \sum_{p=1}^{P-1} h_p(u) \, du \right)\text{,}
\end{equation}
and the cumulative incidence function (CIF) for transitioning from state 0 to any state $p \neq 0$:
\begin{equation}
\Pr_{0 \rightarrow p}(0, t) = \int_0^t S_0(u-) \, h_p(u) \, du, \quad p = 1, \ldots, P-1.
\end{equation}

\begin{figure}[!ht]
\centering
\includegraphics[width=0.5\textwidth]{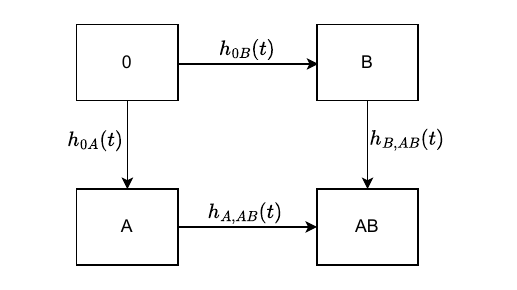}
\caption{A multi-event model describing the joint behavior of A and B.}
\label{fig:multi_event_model}
\end{figure}

\textbf{Multiple events.} In multi-event scenarios, we can observe multiple transitions (events) for each instance $i$, thus the transitions (or event occurrences) are not necessarily mutually exclusive, as in Figure \ref{fig:multi_event_model}.

Such a model can, for example, describe the joint behavior of two events $A$ and $B$; if $h_{0B} = h_{A,AB}$ but $h_{0A} \neq h_{B, AB}$, $A$ is called \emph{locally dependent} on $B$ but $B$ is not locally dependent on $A$. The temporal ordering of events enables an asymmetric notion of dependence, providing more information for causal inference than the symmetric association concepts used in cross-sectional studies.

\section{Survival likelihood}
\label{app:survival_likelihood}

In the standard survival analysis setting, considering a single event of interest, we define the survival time $T_E$ and censoring time $T_C$ as follows:
\[
\begin{array}{c}
T_E = t^{(i)} \ \text{and} \ T_C \geq t^{(i)} \quad \text{if} \ \delta^{(i)} = 1, \\[4pt]
T_E > t^{(i)} \ \text{and} \ T_C = t^{(i)} \quad \text{if} \ \delta^{(i)} = 0.
\end{array}
\]

Combining these cases, the likelihood for the $i$-th observation is:
\begin{equation}
\label{eq:likelihood_function_indep}
\mathcal{L}^{(i)} = \Pr(T_E = t^{(i)}, T_C \geq t^{(i)} \mid x^{(i)})^{\delta^{(i)}} \; \Pr(T_E \geq t^{(i)}, T_C = t^{(i)} \mid x^{(i)})^{1 - \delta^{(i)}}
\end{equation}

Under the assumption of independent censoring~\citep{Emura2018}, we have:
\begin{align}
\mathcal{L}^{(i)} &= \left[ \Pr(T_E = t^{(i)} \mid x^{(i)}) \, \Pr(T_C \geq t^{(i)} \mid x^{(i)}) \right]^{\delta^{(i)}} \left[ \Pr(T_E \geq t^{(i)} \mid x^{(i)}) \, \Pr(T_C = t^{(i)} \mid x^{(i)}) \right]^{1 - \delta^{(i)}} \notag \\
&= \left[ f_E(t^{(i)} \mid x^{(i)}) \, S_C(t^{(i)} \mid x^{(i)}) \right]^{\delta^{(i)}} \left[ S_E(t^{(i)} \mid x^{(i)}) \, f_C(t^{(i)} \mid x^{(i)}) \right]^{1 - \delta^{(i)}} \notag \\
&= \left[ f_E(t^{(i)} \mid x^{(i)})^{\delta^{(i)}} S_E(t^{(i)} \mid x^{(i)})^{1 - \delta^{(i)}} \right] \left[ f_C(t^{(i)} \mid x^{(i)})^{1 - \delta^{(i)}} S_C(t^{(i)} \mid x^{(i)})^{\delta^{(i)}} \right]
\end{align}
where $S_E(t \mid x^{(i)}) = \Pr(T_E \geq t \mid x^{(i)})$, $f_E(t \mid x^{(i)}) = -dS_E(t \mid x^{(i)})/dt$, $S_C(t \mid x^{(i)}) = \Pr(T_C \geq t \mid x^{(i)})$, and $f_C(t \mid x^{(i)}) = -dS_C(t \mid x^{(i)})/dt$.  
Under the non-informative censoring assumption, the term $f_C(t^{(i)} \mid x^{(i)})^{1 - \delta^{(i)}} S_C(t^{(i)} \mid x^{(i)})^{\delta^{(i)}}$ is unrelated to the likelihood for the event times and can simply be ignored~\citep{Emura2018}. Therefore, the likelihood function can be written as:

\begin{equation}\label{eq:likelihood_function_noninfo}
\mathcal{L} = \prod_{i=1}^{N} f_E(t^{(i)} \mid \bm{x}^{(i)})^{\delta^{(i)}} S_E(t^{(i)} \mid \bm{x}^{(i)})^{1 - \delta^{(i)}}, 
\end{equation}
where $f_E(t^{(i)} \mid x^{(i)})$ is the density function of the event time and $S_E(t^{(i)} \mid x^{(i)})$ is the survival function of the event time. Note how the censoring terms $f_C(t^{(i)})$ and $S_C(t^{(i)})$ are absent in \eqref{eq:likelihood_function_noninfo}. If the event is observed, the likelihood contribution is the probability density of the event occurring at that time. If the observation is censored, the likelihood contribution is the probability of surviving beyond that time. The overall likelihood is the product of these contributions across all instances.

\section{Evaluation metrics}
\label{app:evaluation_metrics}

\textbf{Harrell's C-index:} The concordance index (C-index or CI) measures the discriminative performance of a survival model by calculating the proportion of concordant pairs among all comparable pairs. A pair is considered comparable if we can determine who has the event first. It is defined as~\citep{harrell1996}:
\begin{equation}
\text{C-index} = \frac{
\sum_{i,j \in \mathcal{D}} \mathbbm{1}[t^{(i)} < t^{(j)}] \cdot \mathbbm{1}[\eta^{(i)} > \eta^{(j)}] \cdot \delta^{(i)}
}{
\sum_{i,j \in \mathcal{D}} \mathbbm{1}[t^{(i)} < t^{(j)}] \cdot \delta^{(i)}
}
\end{equation}

\noindent where $\eta^{(i)}$ and $\eta^{(j)}$ represent risk scores for individuals $i$ and $j$, respectively.

\textbf{Global C-index:} The global C-index is computed by averaging C-index scores across multiple instances for each event, and was proposed by~\citet{tjandra_hierarchical_2021}. It thus provides an overall measure of a model's discriminative performance in multiple events scenarios. This metric evaluates global consistency, \ie how well does the model discriminate instances \emph{within} an event. It is defined as:
\begin{equation}
\text{Global C-index} = \frac{\sum_{k \in \mathcal{K}}\sum_{i,j \in \mathcal{D}_k} \mathbbm{1}[t^{(i)} < t^{(j)}] \cdot \mathbbm{1}[\eta^{(i)} > \eta^{(j)}] \cdot \delta^{(i)} } 
{\sum_{k \in \mathcal{K}} \sum_{i,j  \in \mathcal{D}_k} 
\mathbbm{1}[t^{(i)} < t^{(j)}] \cdot \delta^{(i)} },
\end{equation}
where $\mathcal{K}$ represents the set of all distinct events, and the index $k$ corresponds to the $k$-th specific event.

\textbf{Local C-index:} The local C-index is computed by averaging C-index scores across multiple events for each instance, and was proposed by~\citet{tjandra_hierarchical_2021}. This metric evaluates local consistency, \ie how well does the model discriminate events \emph{within} an individual. It is defined as:

\begin{equation}
\text{Local C-index} = \frac{\sum_{i \in N}\sum_{k_1, k_2 \in \mathcal{K}^{(i)}} \mathbbm{1}[t_{k_1} < t_{k_2}] \cdot \mathbbm{1}[\eta_{k_1} > \eta_{k_2}] \cdot \delta_{k_1} }
{\sum_{i \in N} \sum_{k_1,k_2 \in \mathcal{K}^{(i)}}
\mathbbm{1}[t_{k_1} < t_{k_2}] \cdot \delta_{k_1}},
\end{equation}
where $\mathcal{K}^{(i)}$ represents the set of all $K$ different events for instance $i$, and $\eta_{k1}$ and $\eta_{k2}$ denote the risk scores associated with the two events, respectively.

\begin{figure}[!ht]
\centering
\includegraphics[width=0.45\columnwidth, trim=10 20 10 20, clip]{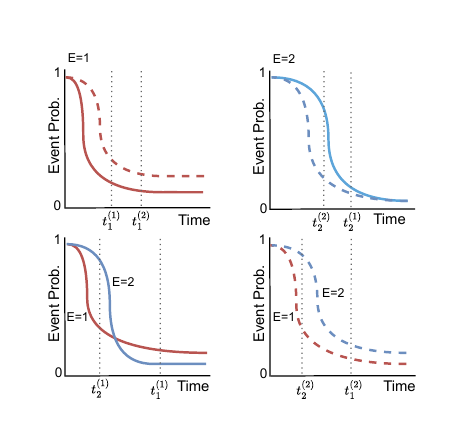}
\caption{Global and local consistency. Red/blue lines represent survival functions $S_{1}(t)$ and $S_{2}(t)$ for event 1 and 2, respectively. Solid/dashed lines represent individuals $i= 1, 2$, respectively. Top-left: globally consistent at $t_1^{(1)}$ and $t_1^{(2)}$ since $S^{(1)}_{1}(t) < S^{(2)}_{2}(t)$ both times. Top-right: globally consistent at $t_2^{(2)}$ and $t_2^{(1)}$ since $S^{(2)}_{2}(t) < S^{(1)}_{2}(t)$ both times. Bottom-left: locally inconsistent at $t_2^{(1)}$ since $S^{(1)}_{2}(t) > S^{(1)}_{1}(t)$ and at $t_1^{(1)}$ since $S^{(1)}_{1}(t) > S^{(1)}_{2}(t)$. Bottom-right: locally inconsistent at $t_2^{(2)}$, since $S^{(2)}_{2}(t) > S^{(2)}_{1}$(t).}
\label{fig:consistency}
\end{figure}

Figure \ref{fig:consistency} shows a visual example of global and local consistency with respect to the survival function. Global consistency is satisfied between instances if the instance-specific curves align with the observed events. Local consistency is satisfied within an instance if the event-specific curves align with the observed events. 

\textbf{BS/IBS:} The Brier Score (BS) is defined as the mean squared difference between the predicted survival curve and the Heaviside step function of the observed event. The integrated Brier Score~\citep{graf1999assessment} (IBS) aggregates the Brier Scores across multiple time points to provide a single measure of model performance. We use inverse probability weighting (IPCW) to handle censored events. The BS is defined as:
\begin{equation}
    \text{BS}(t^*) = 
    \frac{1}{N} \sum_{i \in \mathcal{D}}
      \bigg[\frac{S(t^*\mid{\mathbf{x}^{(i)}})^2 \cdot \mathbbm{1}[t^{(i)} \leq t^*, \delta^{(i)}=1]}{S_C(t^{(i)})}
      + \frac{(1 - S(t^*\mid{\mathbf{x}^{(i)}}))^2 \cdot \mathbbm{1}[t^{(i)} > t^*]}{S_C(t^*)} \bigg],
\end{equation}
\\
which is the mean square error between observed survival status and survival probability at time $t^*$, and $S_C(t^*)$ is the non-censoring probability at time $t^*$. The IBS is defined as:
\begin{equation}
\begin{aligned}     
    \text{IBS} = 
    \frac{1}{N} \sum_{i\in \mathcal{D}}
    \frac{1}{t_{\text{max}}}  \cdot \int_0^{t_{\text{max}}} \text{BS}(t) \, dt,
\end{aligned}
\end{equation}
\\
where $t_{\text{max}}$ is the maximum observed time.

\textbf{MAE:} The mean absolute error (MAE) is the absolute difference between the predicted and actual survival times. Given an individual survival distribution, $S\br{t\mid\bm{x}_{i}} = \text{Pr}\br{T>t\mid\bm{x}_{i}}$, we calculate the predicted survival time $\hat{t}^{(i)}$ using the median time of the predicted survival distribution~\citep{qi_survivaleval_2024}:
\begin{equation}
\hat{t}^{(i)} \ = \ \text{median}\;(S(t\mid\bm{x}_{i})) = S^{-1}(\tau = 0.5\mid\bm{x}_{i})\text{,}
\end{equation}
\begin{equation}
\text{MAE}\;(\hat{t}_{i}, t_{i}, \delta_{i} = 1) \ = \ | \;t_{i}-\hat{t}_{i}\;| \text{.}
\end{equation}

For censored individuals, we calculate the marginal MAE (mMAE), as proposed by~\citet{haider_effective_2020}:
\begin{align}
    & \text{mMAE} = \frac{1}{\sum_{i=1}^{N} \omega^{(i)}} \sum_{i =1}^{N} \omega^{(i)}  \left| [(1 - \delta^{(i)}) \cdot e_{\text{m}}(t^{(i)}) + \delta^{(i)} \cdot t^{(i)} ] - \hat{t}^{(i)} \right| , \notag \\
    & \text{where} \quad    e_{\text{m}}(t^{(i)})=
    \begin{cases}
      t^{(i)} + \frac{\int_{t^{(i)}}^\infty S_{\text{KM}} (t) dt}{S_{\text{KM}} (t)} & \text{if} \quad \delta^{(i)} = 0\\
      t^{(i)} & \text{if} \quad \delta^{(i)} = 1
    \end{cases},       \notag \\ 
    & \text{and} \quad \omega^{(i)} = 
    \begin{cases}
      1 - S_{\text{KM}} (t^{(i)}) & \text{if} \quad \delta^{(i)} = 0\\
      1 & \text{if} \quad \delta^{(i)} = 1
    \end{cases}.
\end{align}

\textbf{D-calibration:} Distribution calibration (D-calibration)~\citep{haider_effective_2020} measures the calibration performance of $S(t)$, expressing to what extent the predicted probabilities can be trusted. We assess this using a Pearson's chi-squared goodness-of-fit test. For any probability interval $[a, b] \in [0, 1]$, we define $D_m(a, b)$ as the group of individuals in the dataset $D$ whose predicted probability of event is in the interval $[a, b]$~\citep{qi_effective_2023}. A model is D-calibrated if the proportion of individuals $|D_m(a, b)|/|D|$ is statistically similar to the amount $b-a$. In the results, we count the number of times a model is D-calibrated (\ie $p>0.05$) for each event over 10 experiments.

\section{Experimental details}
\label{app:experimental_details}

\subsection{Datasets and preprocessing}
\label{app:datasets_and_preprocessing}

Table \ref{tab:datasets} shows an overview of the survival datasets used in this work.

\begin{table}[!ht]
\caption{Overview of the raw datasets. Symbols indicate single-event ($*$), competing-risks ($\dagger$) or multi-event ($\ddagger$) task. Number of events ($K$) is denoted for the respective task. In SEER, "D", "BC" and "HF" denote the death event, death by breast cancer and death by heart failure. In MIMIC-IV, "A", "S", and "D" denote the ARF, shock and death events. In Rotterdam, "R" denotes relapse and "D" denotes death. In PRO-ACT, "SP", "SW", "HA" and "WA" denote the speech, swallowing, handwriting and walking events. In EBMT, "R", "AE", "RAE", "REL" and "D" denote the recovery, adverse events, recovery and adverse events, relapse, and death events.}
\label{tab:datasets}
\centering
\begin{tabular}{@{}lcccccc@{}}
\toprule
Dataset & \makecell{\#Instances\\($N$)} & \makecell{\#Features\\($d$)} & \makecell{\#Events\\($K$)} & \makecell{Max\\$t$} & Event distribution (\%) \\ \midrule
SEER$* \dagger$                 & 19,246    & 17        & \makecell{1/2}   & \makecell{121} & \makecell{D: 45.5\\BC: 12.4, HF: 45.5} \\ \cdashline{1-6}
MIMIC-IV$* \ddagger$    & 24,516    & 100       & \makecell{1/3} & 4,686 & \makecell{D: 37.2\\A: 20.7, S: 19.2, D: 37.2} \\ \cdashline{1-6}
Rotterdam$\dagger \ddagger$     & 2,982     & 10        & 2/3   & 7,043   & \makecell{R: 6.5, D: 50.9\\R: 50.9, D: 42.7} \\ \cdashline{1-6}
PRO-ACT$\ddagger$               & 3,220      & 8         & 4     & 498     & \makecell{SP: 37.8, SW: 31.9,\\HA: 50.3, WA: 60.6} \\ \cdashline{1-6}
EBMT$\ddagger$                  & 2,279     & 6        & 5     & 6229   & \makecell{R: 34.4, AE: 39.8, RAE: 29,\\REL: 16.2, D: 23.4} \\
\bottomrule
\end{tabular}
\end{table}

\textbf{SEER:} The  U.S. Surveillance, Epidemiology, and End Results (SEER) dataset~\citep{ries_cancer_2003} is a comprehensive collection of cancer patient data from approximately 49\% of the U.S. population. It provides information about survival times for cancer patients post-diagnosis. For the single-event case, we consider the outcome of death due to breast cancer, and for the competing-risks case, we consider the outcomes of death due to breast cancer and death due to heart failure, and consider other outcomes as censored. Concerning preprocessing, we select a cohort of 19,246 newly-diagnosed patients (first year of disease) with 17 features, which include demographics and tumor characteristics. We use the raw feature "COD to site recode" to extract the event indicator and apply some special value processing for some of the features, based on the guideline provided by~\citet{wang_survtrace_2022}. For categorical features, missing values are imputed using the mode, followed by one-hot encoding. The data is then standardized to ensure consistency in our analyses. The SEER cohort is available for download at \url{www.seer.cancer.gov}.

\textbf{MIMIC-IV:} The Medical Information Mart for Intensive Care (MIMIC-IV) dataset by \citet{johnson_mimic_2023} comprises electronic health records from patients admitted to intensive care units at the Beth Israel Deaconess Medical Center in the U.S., covering the years from 2008 to 2019. For the single-event case, we consider death since hospital admission and for the multi-event case, we consider the outcomes of acute respiratory failure (ARF), shock and death. Concerning preprocessing, we use the method described by~\citet{gupta2022extensive} to extract 1,672 static features from the dataset, for 26,236 patients whose ages are between 60 and 65 years. To refine our feature set, we apply the unicox feature selection method~\citep{qi2022personalized, simon2003design}, reducing the number of features to the 100 that appear to be the most relevant. In the training set, we assessed the statistical significance of each feature by calculating the p-values using lifelines's~\citep{Davidson-Pilon2019} CoxPHFitter summary. The top 100 features with the smallest p-values were selected. The final set of features is included in the source code. The MIMIC-IV v2.2 is available for download at \url{https://physionet.org/content/mimiciv/2.2/}.

\begin{figure}[!htbp]
\centering
\includegraphics[width=0.3\linewidth, trim=10 10 10 10, clip]{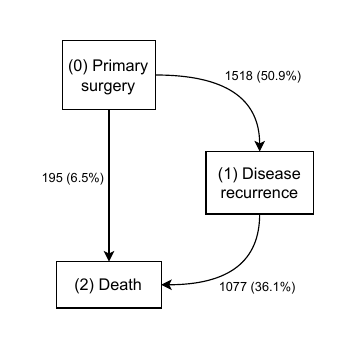}
\caption{Event trajectories in the Rotterdam dataset ($N$=2,982).}
\label{fig:rotterdam_trajectories}
\end{figure}

\textbf{Rotterdam:} The Rotterdam dataset~\citep{royston_external_2013} includes 2,982 primary breast cancer patients from the Rotterdam tumor bank, with 1,546 having node-positive disease. The follow-up time ranged from 1 to 231 months (median 107 months). The dataset has 10 features, which include demographics, tumor characteristics, and treatment information. In the competing-risks case, we consider the outcome of recurrence-free survival time (RFS), which is defined as the time from primary surgery to the earlier of disease recurrence or death from any cause. In the multi-event case, we consider the time to death or disease recurrence, where death censors disease recurrence, but disease recurrence does not censor death. Figure \ref{fig:rotterdam_trajectories} shows the event trajectories in the Rotterdam dataset. Concerning preprocessing, we converted the categorical size groups ('$\leq20$', '$20\text{-}50$', '$>50$') to their approximate median values (10, 35, and 75, respectively). The Rotterdam dataset is available at \url{https://rdrr.io/cran/survival/man/rotterdam.html}.

\textbf{PRO-ACT:} The Pooled Resources Open-Access Clinical Trials (PRO-ACT)\footnote{Data used in the preparation of this article were obtained from the Pooled Resource Open-Access ALS Clinical Trials (PRO-ACT) Database. The data available in the PRO-ACT Database have been volunteered by PRO-ACT Consortium members.} database by \citet{atassi_proact_2014} is the largest ALS dataset in the world, which includes patient demographics, lab and medical records and family history of over 11,600 ALS patients, where ALS disease state is evaluated by the ALS functional rating scale revised (ALSFRS-R)~\citep{Cedarbaum1999}. The assessment contains 12 questions and a maximum score of 48, each representing physical functionality, \eg the ability to walk. Each ALSFRS-R score is between 0 and 4, where 0 represents complete inability with regard to the function, and 4 represents normal function. We use the ALSFRS-R scores to formulate four events: "Speech", "Swallowing", "Handwriting" and "Walking". We consider an event to have occurred if a patient scores 2 or lower on the ALSFRS-R scale during any subsequent follow-up visitation following their baseline visitation. The events can occur in any order and are not mutually exclusive. ALSFRS-R scores are generally monotonic decreasing, but due to the subjectivity of the assessment, scores can occasionally increase post-baseline. We use a maximum follow-up time of 500 days and exclude patients with no recorded history. The PRO-ACT dataset is available for download at \url{https://ncri1.partners.org/ProACT}.

\textbf{EBMT:} The European Society for Blood and Marrow Transplantation (EBMT) dataset~\citep{Rossman2022} consists of 2,279 patient records who underwent blood or marrow transplantation between 1985 and 1998. The dataset records survival times after transplantation for patients suffering from blood cancer. We consider five events in our analysis: "Recovery" (Rec), "Adverse Event" (AE), a combination of both (AE and Rec), "Relapse", and "Death". These events allowed us to evaluate their individual and combined effects on patient prognosis. The dataset includes six features belonging to four baseline covariates: donor-recipient match (where a gender mismatch is defined as a female donor and a male recipient), prophylaxis type, year of transplant, and age at transplant. Concerning preprocessing, we encode the year of transplantation as boolean variables indicating whether the transplant took place during 1990-1994 and 1995-1998, and we classify the age at the transplantation into boolean variables indicating whether the age was less than 20, between 20 and 40, or more than 40 years. The EBMT dataset is available in the PyMSM package at \url{https://github.com/hrossman/pymsm}.

\begin{figure}[!htbp]
\centering
\includegraphics[width=0.5\linewidth, trim=10 10 10 10, clip]{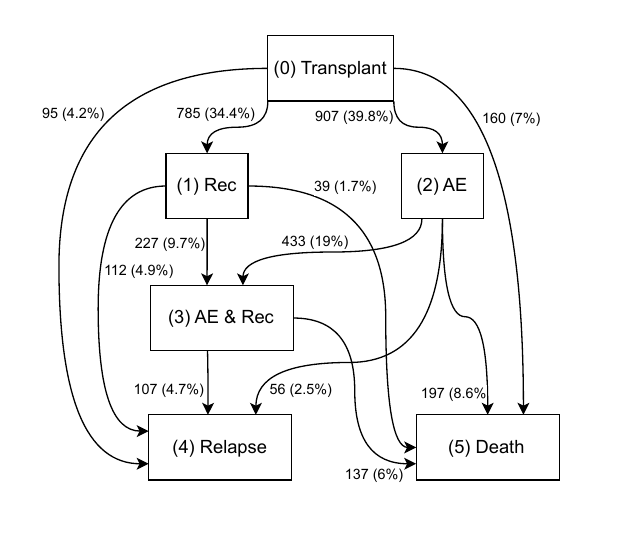}
\caption{Event trajectories in the EBMT dataset ($N$=2,279).}
\label{fig:ebmt_trajectories}
\end{figure}

\subsection{Baselines models}
\label{app:baselines}

\textbf{CoxPH:} The Cox Proportional Hazards model (CoxPH) is a widely used semiparametric model for survival analysis~\citep{cox_regression_1972}. It assumes that the hazard function for an individual $i$ with covariates $\bx^{(i)}$ can be expressed as $h\br{t\vert \bx^{(i)}} = h_0\br{t} \exp \br{f\br{\bm{\bth},\bx^{(i)}}}$, where $h_0(t)$ is an unspecified baseline hazard function, and $f$ is typically a linear function $f(\bm{\bth}, \bx^{(i)}) = \bx^{(i)\top} \bm{\bth}$ producing a risk score. The key assumption of CoxPH is that covariates have a multiplicative effect on the hazard and that this effect is constant over time (\ie proportional hazards). Model parameters $\bm{\bth}$ are estimated by maximizing the Cox partial log-likelihood, which does not require modeling $h_0(t)$ explicitly. The model is implemented from the \emph{scikit-survival} package~\citep{polsterl2020scikit}.

\textbf{CoxNet:} CoxNet is an extension of the Cox Proportional Hazards model that incorporates Elastic Net regularization into the partial log-likelihood optimization~\citep{simon_regularization_2011}. The hazard function retains the same form as in CoxPH, $h\br{t\vert \bx^{(i)}} = h_0\br{t} \exp \br{\bx^{(i)\top} \bm{\bth}}$, but the parameter estimation adds a penalty term $\omega \left[ \alpha \lVert \bm{\bth} \rVert_1 + \frac{1-\alpha}{2} \lVert \bm{\bth} \rVert_2^2 \right]$, where $\omega > 0$ controls the overall strength of regularization, and $\alpha \in [0,1]$ controls the balance between L1 (lasso) and L2 (ridge) penalties. This approach enables feature selection (via L1) while maintaining stability in the presence of multicollinearity (via L2), making it particularly effective in high-dimensional settings. The model is implemented from the \emph{scikit-survival} package~\citep{polsterl2020scikit}.

\textbf{Weibull AFT:} The Weibull Accelerated Failure Time (AFT) model is a parametric survival model in which survival times are assumed to follow a Weibull distribution whose parameters depend on covariates. In the absence of covariates, the scale parameter can be interpreted as the time at which $37\%$ of the population has experienced the event, while the shape parameter determines whether the cumulative hazard is convex or concave, corresponding to accelerating or decelerating hazards. The cumulative hazard function has the closed form $H(t) = (t / \beta)^\eta$, where $\beta > 0$ is the scale parameter and $\eta > 0$ is the shape parameter. When covariates $\bm{x}^{(i)}$ are present, the scale parameter is modeled as $\beta^{(i)} = \exp(-\bm{x}^{(i)\top} \bm{\theta})$ and, optionally, the shape parameter may also depend on $\bm{x}^{(i)}$. Model parameters are estimated by maximizing the full likelihood. The model is implemented from the \emph{lifelines} package~\citep{Davidson-Pilon2019}.

\textbf{GBSA:} Gradient Boosted Survival Analysis (GBSA)~\citep{Ridgeway1999} is a shallow ensemble model that extends gradient boosting methods to survival settings by optimizing a penalized version of the Cox partial log-likelihood. Specifically, it employs an $\ell_2$-norm regularization to prevent overfitting and enhance generalization. Unlike classical gradient boosting, which typically updates a single weak learner or fits all features to the negative gradient, GBSA uses an offset-based approach to flexibly update multiple candidate features at each boosting iteration. This leads to more efficient gradient approximation and model convergence. The model is implemented from the \emph{scikit-survival} package~\citep{polsterl2020scikit}.

\textbf{RSF:} Random Survival Forests (RSF)~\citep{ishwaran_random_2008} adapt the random forest framework to survival analysis by building an ensemble of survival trees, each trained on a bootstrap sample of the data. Each tree recursively partitions the feature space using a survival-specific splitting criterion—typically the log-rank test—to maximize differences in survival outcomes between child nodes. RSF naturally handles nonlinear covariate effects and complex interactions, and it is well-suited for high-dimensional settings without requiring strong parametric assumptions. Censoring is handled natively by the splitting criterion and ensemble aggregation. The model is implemented from the \emph{scikit-survival} package~\citep{polsterl2020scikit}.

\textbf{NFG:} Neural Fine-Gray (NFG)~\citep{jeanselme23a} extends the classical sub-distribution hazard approach of \citet{fine1999proportional} to a flexible neural-network framework for competing risks. The architecture employs a shared embedding network for covariates, followed by separate monotonic neural networks that directly model each instance's CIF, ensuring non-decreasing behavior over time through weight constraints. The CIFs are combined using a risk-balancing mechanism so that their sum remains less than or equal to one, preserving a valid joint probability structure. Training maximizes the exact log-likelihood derived analytically from the CIFs via automatic differentiation, avoiding numerical integration. This formulation allows the model to capture non-linear covariate effects and complex, risk-specific dynamics without assuming a parametric hazard form. The model is implemented from \url{https://github.com/Jeanselme/NeuralFineGray}.

\textbf{DeepSurv:} DeepSurv~\citep{katzman_deepsurv_2018} is a deep learning extension of the Cox Proportional Hazards model that replaces the linear risk function with a nonlinear neural network. The individual hazard function takes the form $h\br{t\vert \bx^{(i)}} = h_0\br{t} \exp \br{f\br{\bth,\bx^{(i)}}}$, where $f(\bth, \bx^{(i)})$ is a feedforward neural network that maps covariates to a continuous risk score. The model is trained by numerically maximizing the Cox partial log-likelihood with respect to the network parameters $\bth$. By capturing nonlinear interactions among covariates, DeepSurv can improve predictive performance over traditional Cox models in complex data settings. The model is implemented from \url{https://github.com/shi-ang/BNN-ISD/}.

\textbf{DeepHit:} DeepHit~\citep{lee_deephit_2018} is a deep discrete-time model designed to handle competing risks, where the occurrence of one event prevents the occurrence of another, as it assumes $\sum_{t=1}^{t_{\text{max}}} \sum_{k=1}^{K} p_k[t] = 1$, where $t_{\text{max}} \in \mathbb{N}_{+}$ is the number of discrete time bins (the maximum discretized time index) in the prediction horizon and $K \in \mathbb{N}_{+}$ is number of competing events (causes). DeepHit contains a shared covariate layer among all events and cause-specific subnetworks for each competing event. The model outputs a probability distribution for each discrete time bin and each competing event via a softmax layer. This architecture allows DeepHit to estimate the joint distribution of survival times and events directly without making assumptions about the underlying stochastic process. It accounts for dependencies among events in $\bth$ but not for individuals who experience no event by time $T$. The model is implemented from \url{https://github.com/chl8856/DeepHit}.

\textbf{Hierarchical:} The hierarchical model~\citep{tjandra_hierarchical_2021} is a deep discrete-time survival model, designed to handle multiple events by predicting event probabilities across multiple temporal granularities. It begins with coarse-scale predictions (\eg monthly) and progressively refines them into finer ones (\eg daily), effectively decomposing long-horizon tasks into simpler intermediate steps. Formally, the model is a multi-task network that takes input \( x \) and produces \( \hat{P} \), a set of event-specific probability distributions \( \hat{p}_k \), computed hierarchically by subnetworks \( \phi_k \). At granularity \( m \), the probability \( \hat{p}_{km} \) is derived from both \( \psi_{km}(\Phi(x)) \) and the preceding level \( \hat{p}_{k,m-1} \), enforcing top-down consistency. During training, a composite likelihood with ranking penalties provides supervision across the event horizon, optimized only for discriminative performance using Harrell's C-index~\citep{harrell1996}. The model is implemented from \url{https://github.com/MLD3/Hierarchical_Survival_Analysis}.

\textbf{MTLR:} The Multi-Task Logistic Regression (MTLR) model~\citep{NIPS2011_1019c809} is a discrete-time survival model that estimates the full survival distribution by training a sequence of dependent logistic regression models—one for each discrete time interval. Unlike models that rely on hazard functions, MTLR directly estimates the probability of survival at each time step, making it flexible and naturally suited for handling censored data. The model captures temporal dependencies between intervals by sharing parameters across the logistic regressors. MTLR has been extended to the competing-risk setting as Deep-CR MTLR~\citep{kim_deep-cr_2021}, which replaces the logistic regressors with neural networks to model nonlinear effects and inter-event dependencies. MTLR is implemented from \url{https://github.com/shi-ang/BNN-ISD/} and Deep-CR MTLR is implemented from \url{https://github.com/bhklab/aaai21_survival_prediction}.

\textbf{DSM:} Deep Survival Machines (DSM)~\citep{nagpal_deep_2021} is a deep continuous-time survival model that estimates the survival function as a weighted mixture of $k$ parametric survival distributions (\eg Weibull or LogNormal). Both the parameters of each distribution and their associated mixture weights are modeled as nonlinear functions of the input covariates, using a shared multilayer perceptron (MLP). This mixture-based formulation allows DSM to capture heterogeneity in the event-time distributions. The model is trained by maximizing the likelihood of observed and censored data. The model is implemented from the \emph{auton-survival} package~\citep{nagpal2022auton}.

\subsection{Hyperparameter settings}
\label{app:hyperparameter_settings}

Tables \ref{tab:coxph_hyperparameters}-\ref{tab:mensa_hyperparameters} list the hyperparameter settings used for each model across all datasets. These values were selected via Bayesian optimization~\citep{snoek_practical_2012} over ten iterations, adopting the hyperparameters leading to the highest Harrell's C-index in the validation set (seed = 0). When applicable, early stopping was used based on the validation loss. Note that the default configuration is used to evaluate computational complexity only.

\begin{table*}[!ht]
\caption{Hyperparameters for CoxPH.}
\label{tab:coxph_hyperparameters}
\centering
\resizebox{0.6\textwidth}{!}{%
\begin{tabular}{lcccccc}
\toprule
Hyperparameter & Default & SEER & Rotterdam & MIMIC-IV & PRO-ACT & EBMT \\
\midrule
Ties method           & breslow & breslow & breslow & breslow & breslow & breslow \\
Regularization & 0       & 10      & 10      & 0.01     & 0       & 0.01 \\
Iterations & 100     & 200     & 100     & 500     & 500     & 100 \\
Tolerance       & 1e-9    & 1e-5    & 1e-5    & 1e-7    & 1e-9    & 1e-5 \\
\bottomrule
\end{tabular}
}
\end{table*}

\begin{table*}[!ht]
\caption{Hyperparameters for CoxNet.}
\label{tab:coxnet_hyperparameters}
\centering
\resizebox{0.6\textwidth}{!}{%
\begin{tabular}{lcccccc}
\toprule
Hyperparameter & Default & SEER & Rotterdam & MIMIC-IV & PRO-ACT & EBMT \\
\midrule
Number of alphas & 100 & 150 & 150 & 150 & 50 & 150 \\
Alpha min ratio & auto & 0.0001 & 0.0001 & 0.0001 & 0.001 & 0.0001 \\
L1 ratio & 0.5 & 0.1 & 1 & 0.75 & 1 & 0.25 \\
Tolerance & 1e-7 & 1e-05 & 1e-7 & 1e-06 & 1e-7 & 1e-05 \\
Max iterations & 100000 & 100000 & 200000 & 100000 & 200000 & 200000 \\
\bottomrule
\end{tabular}
}
\end{table*}

\begin{table*}[!ht]
\caption{Hyperparameters for Weibull AFT.}
\label{tab:weibull_aft_hyperparameters}
\centering
\resizebox{0.6\textwidth}{!}{%
\begin{tabular}{lcccccc}
\toprule
Hyperparameter & Default & SEER & Rotterdam & MIMIC-IV & PRO-ACT & EBMT \\
\midrule
Penalizer  & 0       & 0.001  & 0.0001 & 0.1    & 0.001  & 0 \\
L1 Ratio   & 0       & 1      & 1      & 0      & 0.5    & 0 \\
\bottomrule
\end{tabular}
}
\end{table*}

\begin{table*}[!ht]
\centering
\caption{Hyperparameters for GBSA.}
\label{tab:gbsa_hyperparameters}
\resizebox{0.6\textwidth}{!}{%
\begin{tabular}{lcccccc}
\toprule
Hyperparameter & Default & SEER & Rotterdam & MIMIC-IV & PRO-ACT & EBMT \\
\midrule
Number of estimators & 100 & 50 & 800 & 800 & 800 & 50 \\
Learning rate & 0.1 & 1 & 1 & 1 & 1 & 1 \\
Max depth & 1 & 10 & 1 & 10 & 10 & 3 \\
Min samples split & 10 & 2 & 5 & 10 & 10 & 2 \\
Min samples leaf & 5 & 10 & 5 & 1 & 10 & 1 \\
Max features & \texttt{sqrt} & \texttt{sqrt} & None & \texttt{sqrt} & None & None \\
Dropout rate & 0.0 & 0.5 & 0.5 & 0.5 & 0.5 & 0.25 \\
Subsample & 0.8 & 1.0 & 1.0 & 0.25 & 0.25 & 0.25 \\
\bottomrule
\end{tabular}
}
\end{table*}

\begin{table*}[!ht]
\caption{Hyperparameters for RSF.}
\label{tab:rsf_hyperparameters}
\centering
\resizebox{0.6\textwidth}{!}{%
\begin{tabular}{lcccccc}
\toprule
Hyperparameter & Default & SEER & Rotterdam & MIMIC-IV & PRO-ACT & EBMT \\
\midrule
Number of estimators & 100 & 500 & 200 & 200 & 500 & 1000 \\
Max depth & 1 & 10 & 5 & 5 & 5 & 10 \\
Min samples split & 10 & 2 & 5 & 10 & 10 & 2 \\
Min samples leaf & 5 & 10 & 5 & 1 & 10 & 1 \\
Max features & \texttt{sqrt} & \texttt{sqrt} & \texttt{log2} & \texttt{log2} & None & \texttt{log2} \\
\bottomrule
\end{tabular}
}
\end{table*}

\begin{table*}[!ht]
\caption{Hyperparameters for NFG.}
\label{tab:nfg_hyperparameters}
\centering
\resizebox{0.6\textwidth}{!}{%
\begin{tabular}{lcccccc}
\toprule
Hyperparameter & Default & SEER & Rotterdam & MIMIC-IV & PRO-ACT & EBMT \\
\midrule
Layers & [32] & [64] & [128] & -- & -- & -- \\
Activation & ReLU & ReLU & ReLU & -- & -- & -- \\
Survival layers & [32] & [32] & [32] & -- & -- & -- \\
Dropout & 0.0 & 0.25 & 0.5 & -- & -- & -- \\
Optimizer & Adam & Adam & Adam & -- & -- & -- \\
Multihead & True & True & True & -- & -- & -- \\
\bottomrule
\end{tabular}
}
\end{table*}

\begin{table*}[!ht]
\centering
\caption{Hyperparameters for DeepSurv.}
\label{tab:deepsurv_hyperparameters}
\resizebox{0.6\textwidth}{!}{%
\begin{tabular}{lcccccc}
\toprule
Hyperparameter & Default & SEER & Rotterdam & MIMIC-IV & PRO-ACT & EBMT \\
\midrule
Hidden size              & 32 & 128 & 64  & 32  & 32  & 64 \\
Learning rate            & 0.001 & 0.005 & 0.001 & 0.0005 & 0.005 & 0.005 \\
Weight decay             & 0.01 & 0.01  & 0.1  & 0.01 & 0    & 0.1 \\
Dropout rate             & 0.25 & 0     & 0.5  & 0    & 0    & 0.5 \\
Epochs                   & 1000 & 2000 & 2000 & 1000 & 2000 & 500 \\
Early stopping           & True & True & True & True & True & True \\
Patience                 & 10   & 10   & 20   & 10   & 20   & 20 \\
\bottomrule
\end{tabular}
}
\end{table*}

\begin{table*}[!ht]
\centering
\caption{Hyperparameters for DeepHit.}
\label{tab:deephit_hyperparameters}
\resizebox{0.6\textwidth}{!}{%
\begin{tabular}{lcccccc}
\toprule
Hyperparameter & Default & SEER & Rotterdam & MIMIC-IV & PRO-ACT & EBMT \\
\midrule
Shared layer nodes         & 32   & 16   & 16   & 32   & 64   & 64 \\
Individual layer nodes     & 32   & 64   & 64   & 64   & 128  & 128 \\
Dropout rate               & 0.25   & 0.25   & 0.25   & 0.5    & 0.1    & 0.1 \\
Learning rate              & 0.001  & 0.0001 & 0.005  & 0.0001 & 0.0005 & 0.0005 \\
Weight decay               & 0.01   & 0.1    & 0      & 0.1    & 0      & 0.01 \\
Alpha                      & 0.2    & 0.4    & 0.4    & 0      & 0      & 0.8 \\
Sigma                      & 0.1    & 0.5    & 0.05   & 0.1    & 0.1    & 0.1 \\
Eta multiplier             & 0.8    & 1.0    & 1.0    & 1.0    & 0.5    & 1.0 \\
\bottomrule
\end{tabular}
}
\end{table*}

\begin{table*}[!ht]
\centering
\caption{Hyperparameters for Hierarchical.}
\label{tab:hierarch_hyperparameters}
\resizebox{0.6\textwidth}{!}{%
\begin{tabular}{lcccccc}
\toprule
Hyperparameter & Default & SEER & Rotterdam & MIMIC-IV & PRO-ACT & EBMT \\
\midrule
Theta layer size             & [32] & [50]  & [200] & [200] & [200] & [50] \\
Fine bin layer size          & [32, 5] & [50, 5] & [100, 10] & [150, 10] & [150, 10] & [100, 10] \\
Learning rate                & 0.001 & 0.0005 & 0.001 & 0.001 & 0.001 & 0.001 \\
Regularization constant      & 0.01  & 0.01      & 0.01   & 0.01   & 0.01  & 0.01 \\
Alpha                        & 0.0001 & 1e-5  & 1e-5  & 0     & 0.0001 & 0.001 \\
Sigma                        & 10    & 10    & 5     & 20    & 5     & 5 \\
\bottomrule
\end{tabular}
}
\end{table*}

\begin{table*}[!ht]
\caption{Hyperparameters for MTLR.}
\label{tab:mtlr_hyperparameters}
\centering
\resizebox{0.6\textwidth}{!}{%
\begin{tabular}{lcccccc}
\toprule
Hyperparameter & Default & SEER & Rotterdam & MIMIC-IV & PRO-ACT & EBMT \\
\midrule
Learning rate             & 0.001  & 0.0001 & 0.001  & 0.0001 & 0.0001 & 0.0005 \\
Weight decay              & 0.01   & 0.001  & 0.001  & 0.1    & 0      & 0.1 \\
Dropout rate              & 0.25   & 0      & 0      & 0.1    & 0.1    & 0.25 \\
\bottomrule
\end{tabular}
}
\end{table*}

\begin{table*}[!ht]
\centering
\caption{Hyperparameters for DSM.}
\label{tab:dsm_hyperparameters}
\resizebox{0.6\textwidth}{!}{%
\begin{tabular}{lcccccc}
\toprule
Hyperparameter & Default & SEER & Rotterdam & MIMIC-IV & PRO-ACT & EBMT \\
\midrule
Hidden layers            & 32  & 32  & 32  & 64  & 64  & 64 \\
Learning rate            & 0.001 & 0.005 & 0.005 & 0.005 & 0.005 & 0.001 \\
Iterations               & 10000 & 20000 & 10000 & 10000 & 10000 & 10000 \\
Batch size               & 32    & 32    & 64    & 32    & 64    & 64 \\
Number of dists.         & 3    & 5     & 1     & 5     & 5     & 1 \\
\bottomrule
\end{tabular}
}
\end{table*}

\begin{table*}[!htbp]
\centering
\caption{Hyperparameters for MENSA.}
\label{tab:mensa_hyperparameters}
\resizebox{0.6\textwidth}{!}{%
\begin{tabular}{lcccccc}
\toprule
Hyperparameter & Default & SEER & Rotterdam & MIMIC-IV & PRO-ACT & EBMT  \\
\cmidrule(lr){1-7}
Hidden nodes & 32 & 128 & 32 & 32 & 16 & 128 \\
Learning rate & 0.001 & 0.0005 & 0.001 & 0.001 & 0.0001 & 0.001 \\
Weight decay  & 0.01  & 0.005  & 0.001 & 0.0001 & 0.001 & 0.0001 \\
Dropout rate & 0.25 & 0 & 0.25 & 0.25 & 0.1 & 0.1 \\
Batch size & 32 & 128 & 32 & 64 & 32 & 32 \\
Number of distributions ($\Psi$) & 1 & 3 & 3 & 1 & 3 & 3 \\
Weighting parameter ($\lambda$) & - & - & 0.25 & - & - & 0.5 \\
\bottomrule
\end{tabular}
}
\end{table*}

\subsection{Implementation details}
\label{app:implementation_details}

To train MENSA, we optimize the joint likelihood of event times and event trajectories using stochastic gradient descent (SGD). The algorithm operates over mini-batches of survival data, computing both the multi-event likelihood and the trajectory likelihood at each iteration, and updating the shared and event-specific parameters accordingly. Algorithm \ref{alg:mensa} summarizes the training procedure, including initialization, loss computation, and parameter updates over the training epochs. All experiments were run in Python 3.9 with PyTorch 1.13.1, NumPy 1.24.3, and Pandas 1.5.3 on a workstation with an Intel Core i9-10980XE (3.00 GHz), 64 GB RAM, and an NVIDIA RTX 3090 GPU (CUDA 11.7). All datasets are publicly available, and the code for loading and preprocessing the datasets is provided in the source code repository.

\begin{algorithm}[!htbp]
\caption{Training algorithm for MENSA.}
\label{alg:mensa}
\begin{algorithmic}[1]
    \STATE \textbf{Input:} A survival dataset $\mathcal{D}$ of the form $\{(X^{(i)}, T_\text{obs}^{(i)}, \delta^{(i)})\}_{i=1}^N$; 
           $\mathcal{M}$: the model with initialized parameters \{$\Phi_\theta, \tilde{\beta}, \tilde{\eta}, \zeta, \xi, g_\theta$\}; 
           $\mathcal{T}$: list of trajectories; 
           $\kappa$: learning rate; 
           $L$: number of training epochs; 
           $\lambda$: weighting hyperparameter.
    \STATE \textbf{Result:} $\mathcal{M} = \{\hat{\Phi}_\theta, \hat{\tilde{\beta}}, \hat{\tilde{\eta}}, \hat{\zeta}, \hat{\xi}, \hat{g}_\theta\}$: trained model parameters.
    \STATE \textbf{Initialization:} 
    $\mathcal{M} \gets \texttt{Instantiate}(\mathcal{M})$
    \FOR{$i = 1, \ldots, L$}
        \STATE $[\log f(T_\text{obs}), \log S(T_\text{obs})] \gets \mathcal{M}(\mathcal{D})$ \hfill // $\log f$ and $\log S$ are $N \times P$
        \STATE $\mathcal{L}_{\text{ME}} \gets -\!\!\sum_{p \in \mathcal{P}} w_p \big[\delta_p \log f_p(T_\text{obs}) + (1-\delta_p)\log S_p(T_\text{obs})\big]$ \hfill // Multi-event loss
        \STATE $\mathcal{L}_{\text{trajectory}} \gets 0$
        \FOR{each pair $(A, B) \in \mathcal{T}$}
            \STATE $\mathcal{L}_{\text{trajectory}} \mathrel{+}= \sum_i \delta_A^{(i)} \delta_B^{(i)} \big[-\log S_B(T_A^{(i)})\big]$ \hfill // Accumulate trajectory loss
        \ENDFOR
        \STATE $\mathcal{L}_{\text{total}} \gets (1 - \lambda)\frac{1}{N}\mathcal{L}_{\text{ME}} + \lambda\frac{1}{N}\mathcal{L}_{\text{trajectory}}$ \hfill // Weighted total loss
        \STATE $\{\hat{\Phi}_{\theta}, \hat{\tilde{\beta}}, \hat{\tilde{\eta}}, \hat{\xi}, \hat{\zeta}, \hat{g}_\theta\}^{(i)} \gets \text{SGD}\!\left(\{\cdot\}^{(i-1)}, \nabla \mathcal{L}_{\text{total}}, \kappa \right)$ \hfill // Update step
    \ENDFOR
    \RETURN $\mathcal{M}$
\end{algorithmic}
\end{algorithm}

\section{Statistical and practical significance}
\label{app:stat_tests}

To assess whether performance differences between MENSA and baseline models were statistically and practically meaningful, we conducted paired tests across all datasets and metrics. For each dataset-metric pair, we compared MENSA to every baseline model using per-seed results averaged across events.

\textbf{Statistical testing.} We performed two-sided paired $t$-tests using the per-seed metric values, treating runs with the same random seed as paired observations. The resulting $p$-values were adjusted for multiple comparisons using the Holm correction method ($\alpha=0.05$). Models whose adjusted $p$-values fell below the threshold were considered significantly different from MENSA. The correction was applied independently for each dataset–metric combination.

\textbf{Practical significance.} In addition to statistical testing, we required a minimum effect size of interest (MEI) to ensure that reported differences were not only statistically significant but also practically meaningful. For all discrimination (CI\textsubscript{G}, CI\textsubscript{L}, AUC) and calibration (IBS, mMAE) metrics, the MEI threshold was set to $1.0$ point on the 0-100 scale used for evaluation. For metrics where higher is better (C-indices and AUC), a baseline was considered practically worse if its mean was at least 1.0 point lower than MENSA's mean. For metrics where lower is better (IBS, mMAE), the baseline had to exceed MENSA's mean by at least 1.0 point to be considered practically worse.

\textbf{Reporting.} A dagger symbol ($\dagger$) in the tables indicates models that were both statistically significant (Holm-corrected $p<0.05$) and practically worse than MENSA according to the MEI criterion. D-calibration was excluded from dagger marking since it is reported as a count of calibrated experiments rather than a continuous measure. The full analysis pipeline -- including scaling conventions, dataset ordering, and Holm-adjusted significance checks -- is implemented in a source code repository.

\section{Additional results}
\label{app:additional_results}

\subsection{Single-event results}
\label{app:single_event_results}

Table \ref{tab:single_event_results} shows the discrimination and calibration results on two single-event datasets. In the SEER dataset, MENSA demonstrates solid discriminative ability, ranking third in Harrell’s C-index and second in AUC, while also achieving the second-best IBS and mMAE scores, which underscores its accuracy in survival prediction. In the MIMIC-IV dataset, MENSA shows similarly strong performance, obtaining the second-best Harrell’s C-index, the best overall AUC and IBS, and the lowest mMAE among the deep learning baselines. Taken together, these findings indicate that MENSA is not only competitive with shallow single-event survival methods but also surpasses existing deep learning approaches on several key metrics. This highlights its effectiveness for both risk discrimination and time-to-event prediction in large, single-event datasets.

\begin{table}[!ht]
\centering
\caption{Discrimination and calibration performance (mean $\pm$ SD.) between different models on single-event benchmark datasets, averaged over 10 experiments. For Harrell's CI, AUC, and D-calibration, higher ($\uparrow$) is better. For IBS and mMAE, lower ($\downarrow$) is better. D-calibration counts the number of experiments in which the respective model was D-calibrated.}
\label{tab:single_event_results}
\resizebox{0.7\textwidth}{!}{%
\begin{tabular}{clcccccc}
\toprule
\multirow{2}{*}{\makecell{Dataset}} &
\multirow{2}{*}{\makecell{Model}} &
\multirow{2}{*}{\makecell{Harrell's CI}} &
\multirow{2}{*}{\makecell{AUC}} &
\multirow{2}{*}{\makecell{IBS}} &
\multirow{2}{*}{\makecell{mMAE}} &
\multirow{2}{*}{\makecell{D-Cal}} \\
& \\
\midrule
\multirow{10}{*}{\makecell{SEER\\($K=1$)}}
& CoxPH & 69.8$\pm$0.61 & 74.3$\pm$0.78 & 16.8$\pm$0.30 & 29.9$\pm$0.39 & 4/10 \\
& CoxNet & 69.8$\pm$0.62 & 74.3$\pm$0.79 & 16.8$\pm$0.30 & 29.8$\pm$0.37 & 4/10 \\
& Weibull & 69.8$\pm$0.62 & 74.3$\pm$0.78 & 16.9$\pm$0.30 & 34.3$\pm$0.82 & 10/10 \\
& GBSA & 77.1$\pm$0.53 & 81.7$\pm$0.62 & 15.1$\pm$0.28 & 25.6$\pm$0.33 & 1/10 \\
& RSF & 77.1$\pm$0.55 & 82.1$\pm$0.69 & 14.2$\pm$0.29 & 29.1$\pm$0.62 & 8/10 \\
& MTLR & 70.5$\pm$0.58 & 75.7$\pm$0.81 & 16.4$\pm$0.30 & 29.9$\pm$0.70 & 10/10 \\
& DeepSurv & 76.0$\pm$0.70 & 80.7$\pm$0.78 & 14.8$\pm$0.35 & 29.6$\pm$0.84 & 0/10 \\
& DeepHit & 75.8$\pm$0.79 & 80.8$\pm$0.93 & 14.8$\pm$0.38 & 26.7$\pm$0.46 & 2/10 \\
& DSM & 75.9$\pm$0.93 & 80.8$\pm$1.03 & 14.9$\pm$0.47 & 42.3$\pm$2.31 & 7/10 \\
& MENSA (Ours) & 76.0$\pm$0.65 & 81.0$\pm$0.77 & 14.8$\pm$0.36 & 27.1$\pm$0.71 & 8/10 \\
\cmidrule(lr){1-1}
\multirow{10}{*}{\makecell{MIMIC-IV\\($K=1$)}}
& CoxPH & 74.2$\pm$0.48 & 78.7$\pm$0.60 & 12.9$\pm$0.38 & 5.9$\pm$0.03 & 10/10 \\
& CoxNet & 74.3$\pm$0.48 & 78.8$\pm$0.61 & 12.8$\pm$0.38 & 5.9$\pm$0.04 & 10/10 \\
& Weibull & 74.5$\pm$0.49 & 79.1$\pm$0.63 & 13.0$\pm$0.38 & 5.8$\pm$0.05 & 10/10 \\
& GBSA & 74.9$\pm$0.43 & 79.5$\pm$0.56 & 13.9$\pm$0.39 & 5.9$\pm$0.05 & 8/10 \\
& RSF & 72.5$\pm$0.47 & 78.9$\pm$0.56 & 14.2$\pm$0.40 & 6.0$\pm$0.06 & 8/10 \\
& MTLR & 74.3$\pm$0.52 & 79.5$\pm$0.77 & 12.5$\pm$0.36 & 7.1$\pm$0.24 & 9/10 \\
& DeepSurv & 75.1$\pm$0.43 & 80.0$\pm$0.60 & 12.4$\pm$0.39 & 6.8$\pm$0.10 & 10/10 \\
& DeepHit & 74.5$\pm$0.49 & 79.7$\pm$0.75 & 20.5$\pm$0.57 & 11.7$\pm$0.14 & 0/10 \\
& DSM & 74.4$\pm$0.68 & 79.2$\pm$0.93 & 13.2$\pm$0.44 & 6.4$\pm$0.56 & 5/10 \\
& MENSA (Ours) & 75.4$\pm$0.46 & 80.1$\pm$0.58 & 12.4$\pm$0.34 & 6.1$\pm$0.05 & 4/10 \\
\bottomrule
\end{tabular}%
}
\end{table}

\subsection{Competing-risks results}
\label{app:competing_risks_results}

Table \ref{tab:competing_risks_results} shows the discrimination and calibration results on two competing-risks datasets. In SEER, MENSA performs competitively across all metrics, combining strong discrimination with low prediction error and good calibration. In Rotterdam, MENSA also provides solid discrimination and competitive calibration, though with somewhat higher prediction error compared to the best-performing single-event baselines. However, it has the third-lowest mMAE in this dataset as well. MENSA also achieves good D-calibration results across both datasets. These findings suggest that MENSA extends robustly to competing-risk settings, offering a balanced trade-off between discrimination and calibration performance.

\begin{table*}[!ht]
\centering
\caption{Discrimination and calibration performance (mean $\pm$ SD.) between different models on competing-risks benchmark datasets, averaged over 10 experiments. For global CI, local CI, and AUC, higher ($\uparrow$) is better. For IBS and mMAE, lower ($\downarrow$) is better. D-calibration counts the number of experiments in which the respective model was D-calibrated.}
\label{tab:competing_risks_results}
\resizebox{0.8\textwidth}{!}{%
\begin{tabular}{clcccccc}
\toprule
\multirow{2}{*}{\makecell{Dataset}} &
\multirow{2}{*}{\makecell{Model}} &
\multirow{2}{*}{\makecell{Global CI}} &
\multirow{2}{*}{\makecell{Local CI}} &
\multirow{2}{*}{\makecell{AUC}} &
\multirow{2}{*}{\makecell{IBS}} &
\multirow{2}{*}{\makecell{mMAE}} &
\multirow{2}{*}{\makecell{D-Cal}} \\
& & & & & & & \\
\midrule
\multirow{11}{*}{\makecell{SEER\\($K=2$)}}
& CoxPH & 67.9$\pm$0.61 & 78.4$\pm$0.71 & 66.9$\pm$0.54 & 13.3$\pm$0.18 & 29.0$\pm$0.60 & 14/20 \\
& CoxNet & 68.8$\pm$0.61 & 78.2$\pm$0.60 & 64.8$\pm$0.62 & 13.4$\pm$0.18 & 26.0$\pm$0.71 & 14/20 \\
& Weibull & 67.9$\pm$0.61 & 78.5$\pm$0.71 & 66.8$\pm$0.56 & 13.5$\pm$0.18 & 54.7$\pm$0.94 & 20/20 \\
& GBSA & 74.8$\pm$0.54 & 78.5$\pm$0.70 & 72.7$\pm$0.67 & 12.4$\pm$0.15 & 23.8$\pm$0.68 & 11/20 \\
& RSF & 74.4$\pm$0.56 & 81.0$\pm$0.49 & 73.7$\pm$0.66 & 12.0$\pm$0.16 & 38.8$\pm$1.78 & 18/20 \\
& NFG & 72.8$\pm$0.55 & 80.0$\pm$0.46 & 68.6$\pm$0.61 & 12.6$\pm$0.16 & 185.8$\pm$11.84 & 7/20 \\
& MTLR-CR & 67.4$\pm$0.70 & 73.8$\pm$1.06 & 66.1$\pm$0.82 & 16.0$\pm$0.39 & 40.6$\pm$0.73 & 0/20 \\
& DeepSurv & 73.4$\pm$0.61 & 79.5$\pm$0.56 & 70.3$\pm$0.47 & 12.3$\pm$0.18 & 31.7$\pm$2.54 & 11/20 \\
& DeepHit & 71.5$\pm$0.42 & 78.5$\pm$0.66 & 68.5$\pm$0.58 & 12.7$\pm$0.14 & 135.2$\pm$15.01 & 8/20 \\
& DSM & 73.0$\pm$0.73 & 79.2$\pm$1.00 & 70.1$\pm$0.65 & 12.5$\pm$0.24 & 74.7$\pm$9.09 & 15/20 \\
& Hierarch. & 71.4$\pm$1.00 & 69.8$\pm$1.33 & 68.2$\pm$1.10 & 15.4$\pm$0.40 & 38.0$\pm$1.48 & 0/20 \\
& MENSA (Ours) & 72.9$\pm$0.66 & 78.7$\pm$0.80 & 69.9$\pm$0.66 & 12.4$\pm$0.19 & 26.0$\pm$0.89 & 17/20 \\
\cmidrule(lr){1-1}
\multirow{11}{*}{\makecell{Rotterdam\\($K=2$)}}
& CoxPH & 69.0$\pm$1.49 & 86.9$\pm$1.85 & 78.8$\pm$1.57 & 13.8$\pm$2.71 & 252.2$\pm$84.82 & 20/20 \\
& CoxNet & 69.0$\pm$1.50 & 86.9$\pm$1.79 & 78.9$\pm$1.49 & 13.8$\pm$2.67 & 307.6$\pm$125.96 & 20/20 \\
& Weibull & 69.1$\pm$1.56 & 88.4$\pm$1.49 & 78.9$\pm$1.61 & 13.9$\pm$3.09 & 420.0$\pm$72.95 & 15/20 \\
& GBSA & 75.2$\pm$1.85 & 88.9$\pm$1.53 & 71.8$\pm$2.47 & 14.7$\pm$2.58 & 21.0$\pm$1.30 & 20/20 \\
& RSF & 69.7$\pm$1.49 & 88.6$\pm$1.29 & 78.6$\pm$1.77 & 13.4$\pm$2.35 & 224.4$\pm$58.86 & 20/20 \\
& NFG & 66.2$\pm$3.86 & 89.0$\pm$1.57 & 75.3$\pm$3.80 & 15.1$\pm$3.05 & 810.1$\pm$129.17 & 13/20 \\
& MTLR-CR & 68.4$\pm$1.81 & 86.2$\pm$1.69 & 78.3$\pm$1.73 & 16.0$\pm$3.89 & 54.3$\pm$7.57 & 3/20 \\
& DeepSurv & 69.1$\pm$1.62 & 88.3$\pm$1.93 & 78.7$\pm$1.78 & 13.9$\pm$2.53 & 103.4$\pm$30.50 & 20/20 \\
& DeepHit & 62.6$\pm$2.52 & 88.8$\pm$1.50 & 77.0$\pm$2.32 & 15.1$\pm$2.74 & 33.3$\pm$5.26 & 15/20 \\
& DSM & 69.9$\pm$1.40 & 88.8$\pm$1.37 & 78.2$\pm$2.01 & 13.5$\pm$2.47 & 139.3$\pm$4.09 & 17/20 \\
& Hierarch. & 65.2$\pm$1.72 & 73.4$\pm$4.57 & 50.0$\pm$0.06 & 76.1$\pm$13.31 & 83.4$\pm$7.46 & 0/20 \\
& MENSA (Ours) & 67.3$\pm$1.71 & 88.9$\pm$1.54 & 77.1$\pm$1.84 & 15.2$\pm$5.23 & 48.9$\pm$7.46 & 18/20 \\
\bottomrule
\end{tabular}%
}
\end{table*}

\section{Computational analysis}
\label{app:computational_analysis}

\begin{table}[!ht]
\centering
\caption{Comparison between the proposed model and selected state-of-the-art baselines in terms of space complexity. $d$: number of features, $K$: number of events, $B$: number of discrete time bins, $M$: levels of granularity, $\Psi$: number of mixture distributions, $P$: number of states.}
\label{tab:space_complexity}
\resizebox{0.55\textwidth}{!}{%
\begin{tabular}{lccc}
\toprule
\textbf{Model} & Single-event & Competing risks & Multi-event \\
\midrule
DeepSurv & $\mathcal{O}(d)$ & $K\mathcal{O}(d)$ & $K\mathcal{O}(d)$ \\
DeepHit & $\mathcal{O}(Bd)$ & $\mathcal{O}(KBd)$ & $K\mathcal{O}(Bd)$ \\
MTLR & $\mathcal{O}(Bd)$ & $\mathcal{O}(KBd)$ & $K\mathcal{O}(Bd)$ \\
DSM & $\mathcal{O}(\Psi + d)$ & $\mathcal{O}(K\Psi + d)$ & $K\mathcal{O}(\Psi + d)$ \\
Hierarch. & $\mathcal{O}(MB + d)$ & $\mathcal{O}(MKB + d)$ & $\mathcal{O}(MKB + d)$ \\
MENSA (Ours) & $\mathcal{O}(\Psi + d)$ & $\mathcal{O}(P(\Psi + d) + d)$ & $\mathcal{O}(P(\Psi + d) + d)$ \\
\bottomrule
\end{tabular}
}
\end{table}

Table \ref{tab:space_complexity} compares selected deep learning baselines and MENSA in terms of space complexity. This analysis is conducted in single-event, competing-risks and multi-event scenarios. For a fair comparison, we configure all models with a single hidden layer with 32 nodes. By nature of design, models that cannot learn events jointly have to be trained $K$ times in the multi-event scenario. In this case, the DeepSurv model, as an example, has a $K\mathcal{O}(d)$ space complexity, where $d$ is the number of features, assuming training is not done in parallel. This can lead to significant increases in the number of trainable parameters in high-dimensional datasets with many events, making it a less-attractive solution. MENSA does not suffer from this problem, as it can train on multiple events at once with a space complexity of $\mathcal{O}(P(\Psi + d) + d)$.

\begin{figure}[!ht]
\centering
\includegraphics[width=0.85\columnwidth]{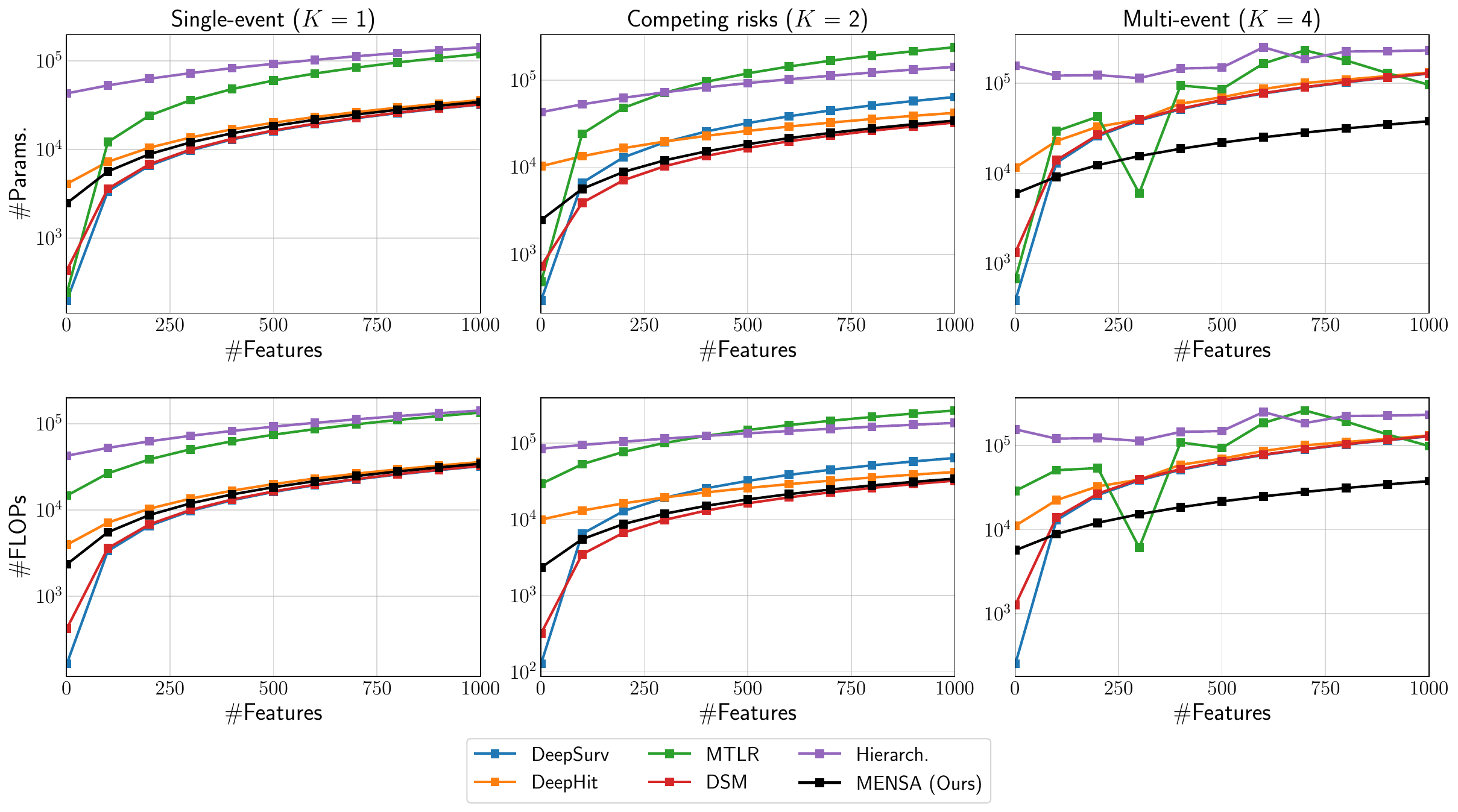}
\caption{Number of features vs. number of trainable parameters (upper row) and number of floating-point operations (lower row) for single-event $(K=1)$, competing-risks $(K=2)$ and multi-event $(K=4)$ scenarios.}
\label{fig:model_complexcity}
\end{figure}

Figure~\ref{fig:model_complexcity} shows the number of trainable parameters and FLOPs (computed using the \texttt{fvcore} library\footnote{Meta Research's fvcore library, available at \url{https://github.com/facebookresearch/fvcore} (September 1, 2025).}) as a function of the number of features. In the single-event case, MENSA uses fewer parameters and FLOPs than Hierarchical, MTLR, and DeepHit, and only slightly more than DeepSurv and DSM. In competing-risks, MENSA remains more efficient than most models except DSM, whose overhead is marginally lower. For datasets with over 100 features, differences are negligible. In the multi-event scenario, baseline models trained separately per event require $K$ trainings, while MENSA trains all events jointly, achieving a notably lower parameter count and FLOPs than all baselines, including the Hierarchical model (see Table \ref{tab:comp_analysis}). Overall, MENSA is the most computationally efficient among deep learning models in multi-event settings.

\begin{table}[!ht]
\centering
\caption{Relative reduction in parameters and FLOPs for MENSA compared to deep learning baselines in the multi-event setting (\(K=4\)) with 100 features.}
\label{tab:comp_analysis}
\resizebox{0.5\linewidth}{!}{
\begin{tabular}{lcc}
\toprule
Model & \#Params. reduction & \#FLOPs reduction \\
\midrule
DeepSurv  & 1.4$\times$ & 1.5$\times$ \\
Deephit   & 2.5$\times$ & 2.5$\times$ \\
MTLR      & 3.2$\times$ & 5.7$\times$ \\
DSM       & 1.5$\times$ & 1.5$\times$ \\
Hierarch.  & 13$\times$  & 13$\times$ \\
\bottomrule
\end{tabular}
}
\end{table}

\section{Proof of non-proportional hazards}
\label{app:proof_of_proportional_hazards}

This section addresses how the proposed method overcomes the limitations associated with the proportional hazards assumption~\citep{cox_regression_1972}. We begin by reviewing how a traditional Weibull distribution adheres to the proportional hazards property and then show, intuitively, why MENSA, which uses a mixture of Weibull distributions, does not meet this assumption.

It is well-established that the Weibull distribution satisfies the proportional hazards assumption when its shape parameter, $\eta$, is independent of the covariates $\bm{x}^{(i)}$~\citep{klein2006survival}. The PDF, CDF, survival function, and hazard function for a Weibull distribution are given by:
\begin{align}
f(t; \beta, \eta) &= \frac{\eta}{\beta} \left( \frac{t}{\beta} \right)^{\eta - 1} \exp\left( -\left( \frac{t}{\beta} \right)^\eta \right), \\
F(t; \beta, \eta) &= \int_{\tau=0}^t f(\tau; \beta, \eta) \, d\tau =  1 - \exp\left( -\left( \frac{t}{\beta} \right)^\eta \right), \\
 S(t; \beta, \eta) &= 1 - F(t;\beta, \eta) = \exp\left( -\left( \frac{t}{\beta} \right)^\eta \right), \\
h(t; \beta, \eta) &= \frac{f(t; \beta, \eta)}{S(t; \beta, \eta)} = \frac{\eta}{\beta} \left( \frac{t}{\beta} \right)^{\eta - 1}. 
\end{align}

To demonstrate that the Weibull distribution satisfies the proportional hazards assumption, we parameterize the scale parameter, $\beta$, as a function of the covariates $\bm{x}^{(i)}$:
\begin{equation} 
\beta(\bm{x}^{(i)}) \ = \ \beta_0 \exp(-\gamma^\top \bm{x}^{(i)}), 
\end{equation}

where $\beta_0$ is the baseline scale parameter, and $\gamma$ is a vector of regression coefficients. Substituting $\beta(\bm{x}^{(i)})$ into the hazard function yields:
\begin{align} 
h(t; \bm{x}^{(i)}) \ 
&= \ \frac{\eta}{\beta(\bm{x}^{(i)})} \left( \frac{t}{\beta(\bm{x}^{(i)})} \right)^{\eta - 1} \\ 
&= \ \frac{\eta}{\beta_0 \exp\left(-\gamma^\top \bm{x}^{(i)}\right)} \left( \frac{t}{\beta_0 \exp\left(-\gamma^\top \bm{x}^{(i)}\right)} \right)^{\eta - 1} \notag \\ 
&= \ \frac{\eta \exp\left(\gamma^\top \bm{x}^{(i)}\right)}{\beta_0} \left( \frac{t \cdot \exp\left(\gamma^\top \bm{x}^{(i)}\right)}{\beta_0} \right)^{\eta - 1} \notag \\ 
&= \ \frac{\eta}{\beta_0^\eta} t^{\eta - 1} \exp\left(\eta \gamma^\top \bm{x}^{(i)}\right). \notag
\end{align}

We define the baseline hazard function as:
\begin{equation} 
\label{eq:baselines_hazard}
h_0(t) \ = \ \frac{\eta}{\beta_0^\eta} \cdot t^{\eta - 1}.
\end{equation}

Therefore, the hazard function can be expressed as:
\begin{equation}
\label{eq:hazard_ph_format}
h(t; \, \bm{x}^{(i)}) \ = \ h_0(t) \exp\left(\eta \gamma^\top \bm{x}^{(i)}\right). \end{equation}

This formulation shows that the Weibull hazard function is the product of a time-dependent baseline hazard, $h_0(t)$, and a time-independent relative hazard, $\exp\left(\eta \gamma^\top \bm{x}^{(i)}\right)$. Hence, when the shape parameter $\eta$ is constant, and only the scale parameter $\beta$ varies with the covariates, the Weibull distribution satisfies the proportional hazards assumption. In contrast, both the Deep Survival Machines (DSM)~\citep{nagpal_deep_2021} and the proposed MENSA model, which utilize a mixture of Weibull distributions, do not satisfy this assumption. This can be attributed to two reasons:

\begin{enumerate}
\item In DSM and MENSA, both the shape parameter $\eta$ and the scale parameter $\beta$ depend on the covariates $\bm{x}^{(i)}$. Substituting an $\bm{x}^{(i)}$-dependent $\eta(\bm{x}^{(i)})$ into the baseline hazard definition prevents the hazard function from being expressed in a proportional hazards form, as shown in \eqref{eq:hazard_ph_format}, thus violating the assumption.
\item Since both DSM and MENSA model the survival distribution as a mixture of Weibulls, their hazard functions can be expressed through a weighted combination of individual Weibull distributions. Thus, given mixture weights, $g_\psi(\bm{x}^{(i)})$, the PDF \eqref{eq:mensa_pdf} and CDF \eqref{eq:mensa_cdf} functions, their hazard function can be described as:
\begin{equation}
h\left(t; \, \beta, \eta, \bm{x}^{(i)}\right) \ = \ \frac{f\left(t; \, \beta, \eta, \bm{x}^{(i)}\right)}{S\left(t; \,\beta, \eta, \bm{x}^{(i)}\right)} 
\ = \ \frac{\sum_{\psi=1}^{\Psi} g_\psi(\bm{x}^{(i)}) f_\psi\left(t; \, \beta_\psi(\bm{x}^{(i)}), \eta_\psi(\bm{x}^{(i)})\right)}{1 - \sum_{\psi=1}^{\Psi} g_\psi(\bm{x}^{(i)}) F_\psi\left(t; \, \beta_\psi(\bm{x}^{(i)}), \eta_\psi(\bm{x}^{(i)})\right)}.
\end{equation}
\end{enumerate}

Because both the shape and scale parameters, as well as the mixture weights, depend on the covariates $\bm{x}^{(i)}$, and the marginal density is modeled as a mixture of Weibull distributions, the resulting hazard function is not proportional across covariates or have a simple linear relationship in $t$ or $\log(t)$. Unlike the traditional Weibull model, mixture-based approaches, such as DSM and MENSA, deviate from the proportional hazards assumption, offering greater flexibility and potentially more accurate representations of patient risk over time.

\end{document}